\documentclass[journal]{IEEEtran}
\ifCLASSINFOpdf
\else
\fi
\usepackage{amsmath,amsfonts}
\usepackage{algorithmic}
\usepackage{algorithm}
\usepackage{array}
\usepackage[caption=false,font=normalsize,labelfont=sf,textfont=sf]{subfig}
\usepackage{textcomp}
\usepackage{stfloats}
\usepackage{url}
\usepackage{verbatim}
\usepackage{graphicx}
\usepackage{cite}
\usepackage{fancybox}
\usepackage{hyperref}

\def\Fevol{\textit{EvolF\/}}  

\begin{document}
\title{Evolution \& Foundation: AI Shares Creative Control}

\author{ Dylan Banarse\thanks{D. Banarse is with Google DeepMind; S. Todd is with London Geometry Ltd.; W. Latham and F. Fol Leymarie are with the School of Computing at Goldsmiths, University of London.}, %
    Stephen Todd, %
    William Latham, and %
    Frederic Fol Leymarie %
    }

\maketitle

\begin{abstract}  
This paper investigates the creative process of automated design and artistic evaluation using an evolutionary system. We consider how a multimodal artificial intelligence (AI) model can communicate and guide a combined generative and evolutionary computational system. This creates a framework for the evolution of aesthetically pleasing complex 3D organic forms by integrating genetic algorithms with the visual reasoning capabilities of large-scale AI foundation models. 

The framework shifts the artist role from that of intensive direct selection to one of system design; transferring detailed step-by-step curation to an AI agent capable of multimodal aesthetic judgement. This framework enables the human artist/designer to rapidly traverse large areas of multi-dimensional evolutionary parameter space to find creative outcomes based on their semantic targets.

Detailed audit trails of the AI's aesthetic reasoning are generated for each experiment. Interactive visualisation tools, together with AI-generated summaries and evolutionary narratives, enable deep exploration into each evolutionary experiment and providing a transparent insight into the AI-guided process.

\end{abstract}

\begin{IEEEkeywords}
computational evolution, genetic algorithm, AI foundation multimodal model, automated curation, evolutionary art, generative AI
\end{IEEEkeywords}

\IEEEpeerreviewmaketitle

\begin{figure*}
\begin{centering}
\shadowbox{\begin{minipage}[t]{0.338\columnwidth}%
\begin{center}
\includegraphics[width=1\columnwidth]{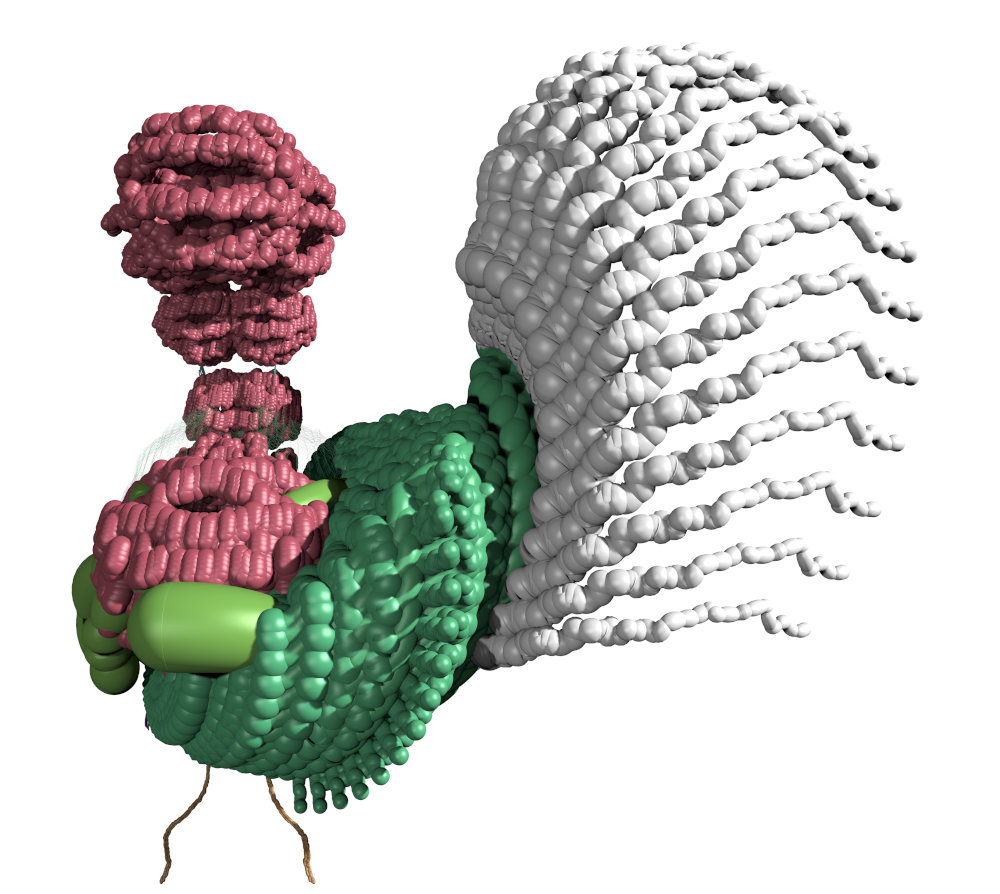}
\par\end{center}
\begin{center}
\includegraphics[width=1\columnwidth]{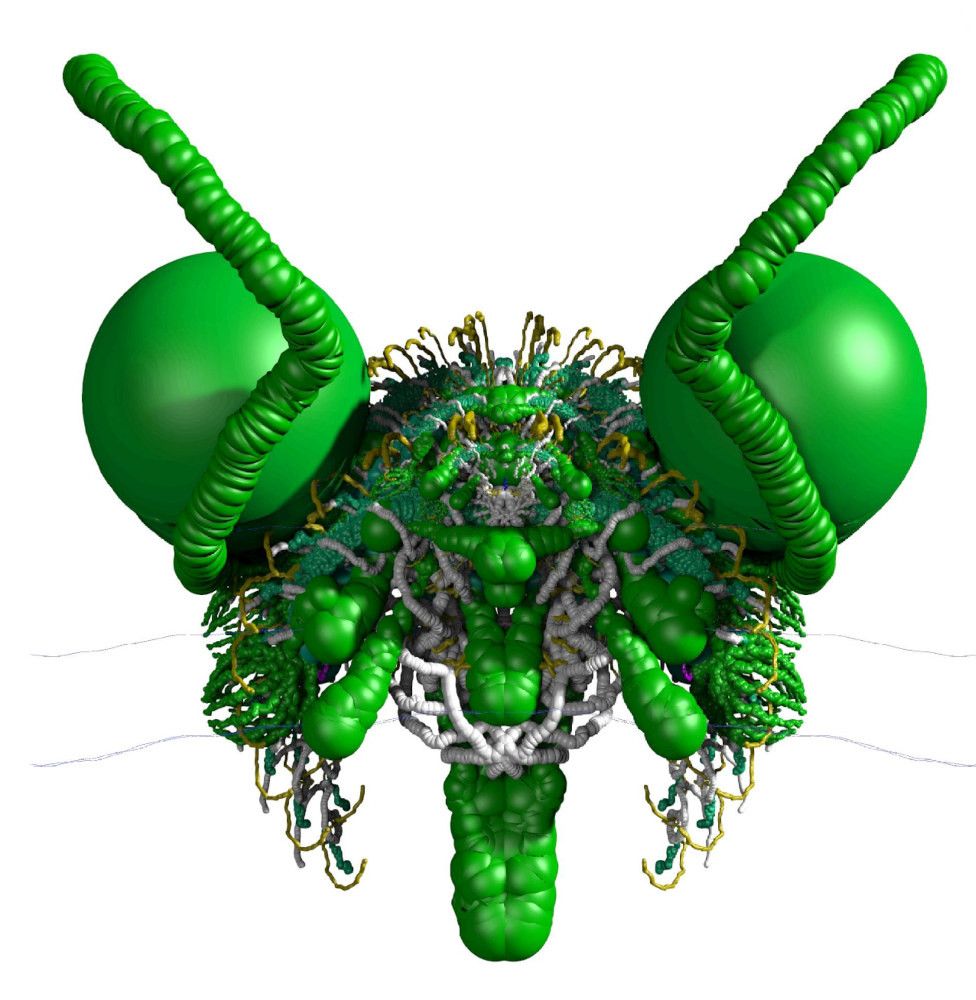}
\par\end{center}
\begin{center}
\includegraphics[width=1\columnwidth]{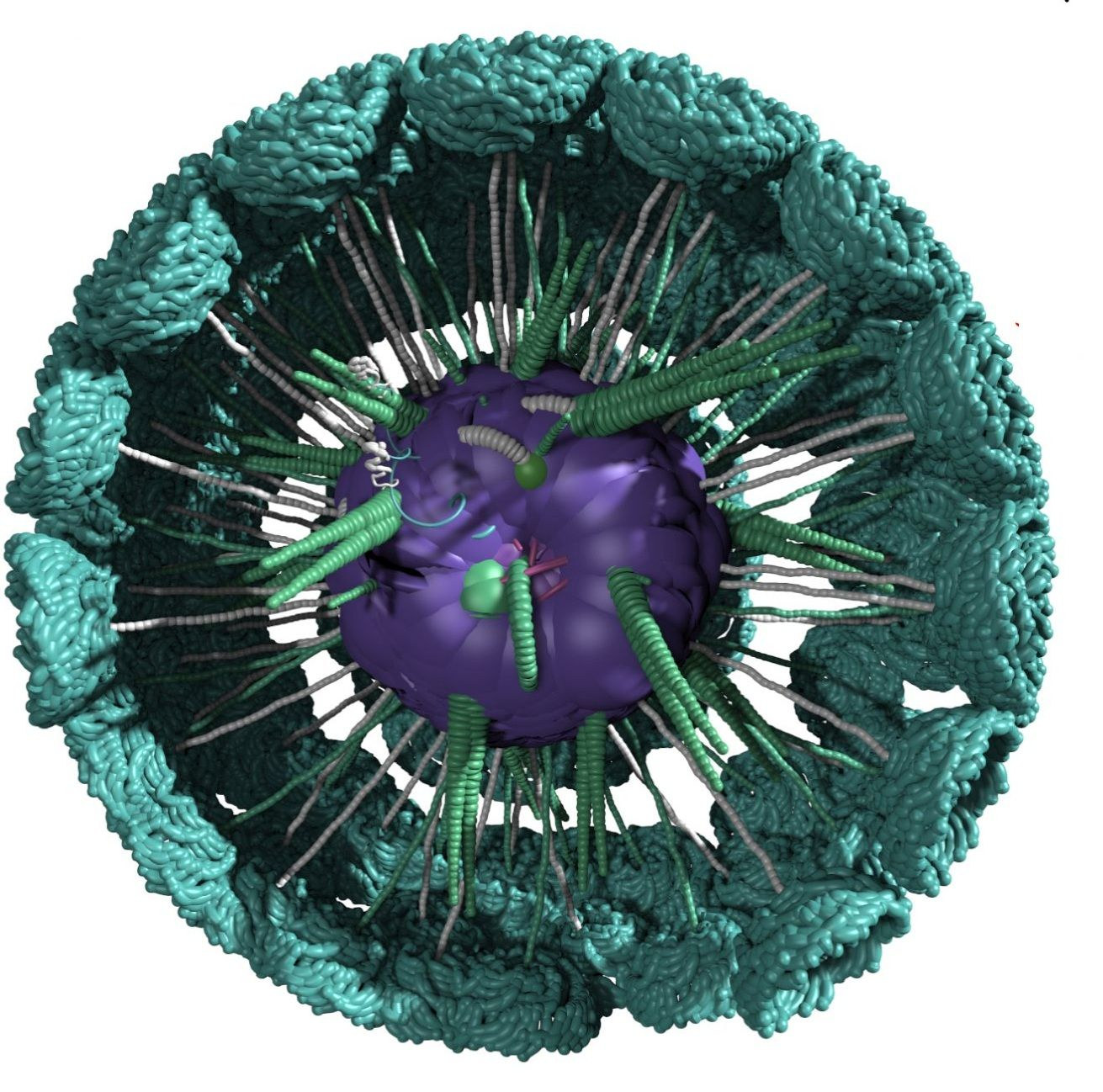}
\par\end{center}%
\end{minipage}}%
\shadowbox{\begin{minipage}[t]{1.58\columnwidth}%
\begin{center}
\includegraphics[width=1\columnwidth]{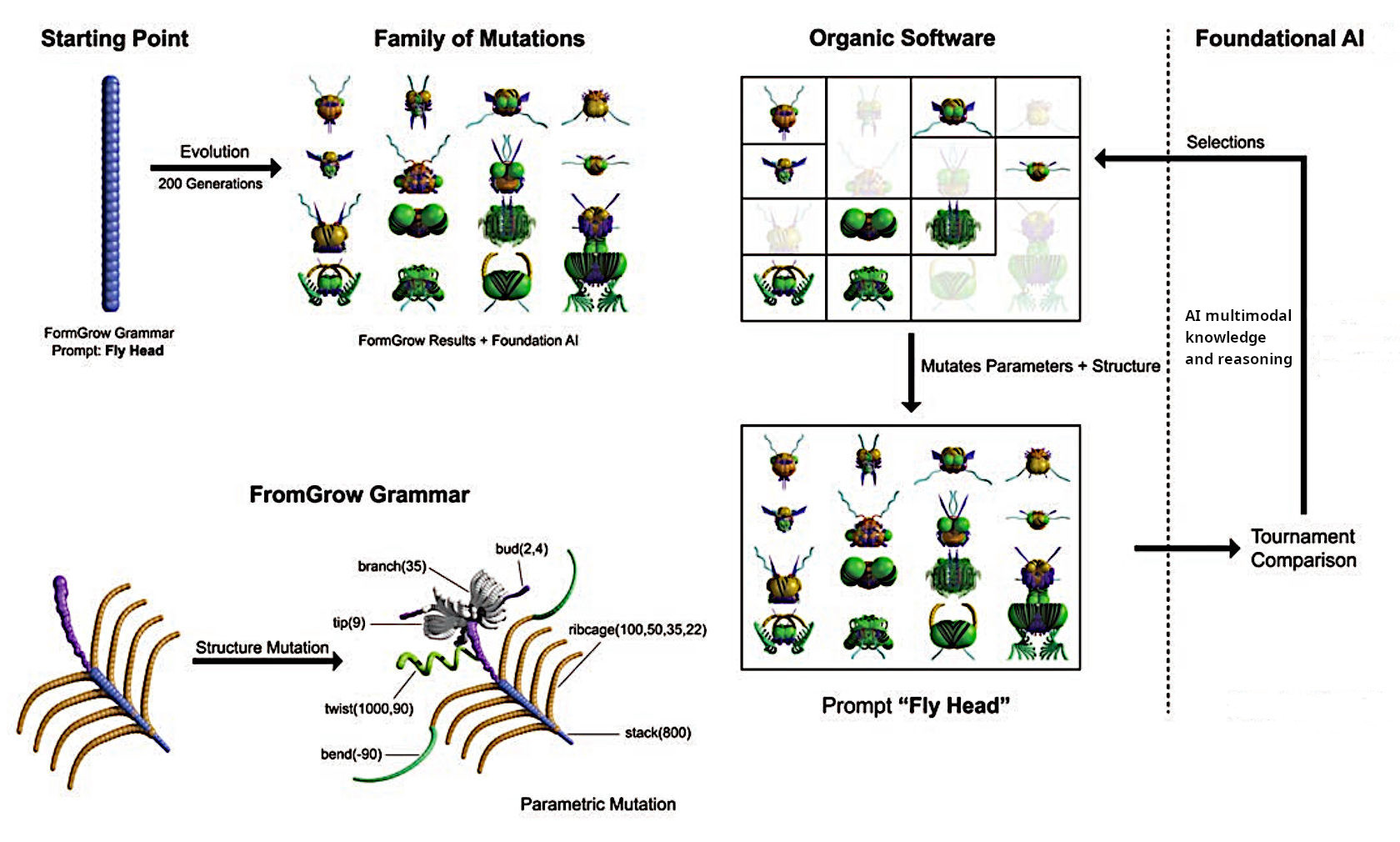}
\par\end{center}
\begin{center}
Evolution-based ``Organic'' working jointly with the selection guidance from a multimodal LLM.
\par\end{center}%
\end{minipage}}
\par\end{centering}
\caption{Left: Three examples of resulting 3D forms produced by our combined evolutionary art and multimodal foundation system: ``chicken'', ``fly head'', ``diatom''. Right: Overview of our Evolution+Foundation system.}
\label{fig:teaser}

\end{figure*}

\section{Introduction}

\IEEEPARstart{T}{he} field of evolutionary art has long been defined by the symbiotic relationship between human aesthetic judgment and algorithmic variation. An already long history before the computer age had flourished, where artists and designers engaged with the algorithmic generation of patterns, such as in musical scores, or architectural rule-based systems as in Palladian aesthetics, or tiling designs such as by M.C. Escher, and more. Computer Evolutionary Art emerged as a discipline in the 1980's \cite{Lambert2013} as the graphical capacities of computers kept increasing.
Initially, the creative process remained essentially tethered to human intervention \textit{for every selective step}. More recently, there has been extensive research into letting the computer be the curator making evolutionary selection, removing the artist from the evolutionary loop. The research is mainly based on the design of aesthetic cost functions \cite{DiPaola2007,denHeijer2010,Machado2021}. This puts extensive weight on the technical designers of the cost functions.

Here we introduce and explore a novel framework termed ``Evolution and Foundation'' (short: \Fevol) which uses an AI multimodal model --- in our case, Google’s Gemini foundation model --- to perform the step-by-step selections as an AI agent capable of assessing abstract visuals against specific semantic goals. By leveraging the multimodal reasoning capabilities of Gemini, we further transition the role of the human artist: from that of a curator using technically crafted algorithms, to that of a designer who can exercise higher level control on the selection process by using suitable AI prompts while judging the quality of final outcomes.

Central to this human-machine collaboration is the concept of \textit{pareidolia} for machines.\footnote{Pareidolia: ``the tendency for perception to impose a meaningful interpretation on a nebulous stimulus, usually visual, so that one detects an object, pattern, or meaning where there is none'' (from \url{https://en.wikipedia.org/wiki/Pareidolia}).} Just as humans find familiar meaning in incoherent shapes, foundation models can be prompted to identify emergent content within the abstract outputs of computerised genetic algorithms. This methodology is tested through a ``litmus test'' experiment: for example, evolving a recognizable ``chicken-like'' archetype (Fig.~\ref{fig:teaser}, top left) from a starting population of abstract ``digital clay''.
By documenting every ``thought'' and tournament decision made by the AI agent, we can provide a narrative based on the machine's aesthetic reasoning, bridging the gap between procedural generative systems and the human-like interpretation of semantic forms.
Concretely, we make three main contributions:
\begin{enumerate}
    \item The Evolution \& Foundation (\Fevol) concept for 3D designs and 3D artistic explorations.
    \item The technical solutions to position bias and creative disruption (Binary tournaments with PixelScore).
    \item The conceptual shift from the artist as a curator to the ``Artist as System Designer Integrating AI into their Evolutionary (Art) Practice''.
\end{enumerate}

\section{Background}
\label{sec:back}

Computer Evolutionary Art emerged in the 1980's, in particular with the pioneering works by the tandem of artist William Latham and mathematician Stephen Todd. This approach treats the artist as a ``gardener'' who selects and breeds digital forms from a near-infinite genetic space \cite{Todd1992}. Early systems like FormGrow, first designed in 1987, utilized a rule-based grammar of primitive shapes and geometric transformations to generate complex organic structures. However, a manual selection process required for each generation was often tedious, leading to the development by Latham and Todd of the Mutator computer system in 1989 to automate the application of changes and present the artist with on-screen visual options for selection \cite{Latham1989,Mutator1991}.

Other early key contributors to the infancy of this discipline include Richard Dawkins and his ``biomorphs'' \cite{Dawkins1986}, Karl Sims for his interactive evolutionary graphics \cite{Sims1991} and early (circa 1994) 3D articulated evolving artificial-life experiments \cite{Sims1994}, Lindenmayer et al. for their approach to generative grammars first developed in biological studies and later applied in computer graphics \cite{AlgoBeauty1990} and afterwards used in artistic explorations, such as by McCormack~\cite{McCormack1993,McCormack2004}.

Good surveys and overviews of the various approaches and key contributors include: Lambert et al. who focus on the initial 1980-1993 period  \cite{Lambert2013}, Bentley et al. who cover works through the 1990's including applications in art, design music, and artificial life (ALife) \cite{Bentley1999,Bentley2001}, Antunes et al. who consider the application of evolutionary art at the level of ecosystems of agents exchanging information or tokens of items like materials or energy \cite{Antunes2014}.
Whitelaw considers the history until the early 2000's of generative and evolutionary art and its application to ALife from a media-theory viewpoint \cite{Whitelaw2004}. Lai et al. give a recent review of methods to procedurally generate virtual ``creatures'': characters with human or animalistic forms and articulations, ready for animations \cite{Lai2021}. A recent survey by Wu et al. focuses on the potentials of having evolutionary algorithms (EAs) and LLMs be combined in various ways to benefit from their complementary features, such as augmenting the search capabilities of LLMs, or improving EAs in convergence towards useful solutions \cite{Wu2025}.

Closer to our work and concerns is the topic of evolution driven by aesthetics, such as explored in \cite{DiPaola2007,denHeijer2010,Wang2020, Machado2021}. More recently, McCormack presents a case study based on his earlier work from 2001, ``Eden'' \cite{Antunes2014}, to explore the potential of mixing current generative AI methods with evolutionary art \cite{McCormack2024}.

A significant advance towards AI-based aesthetics was the development of the foundation model CLIP by Radford et al. \cite{pmlr-v139-radford21a}; a pre-trained model with shared embeddings for 400 million (image, text) pairs. Due to the shared embeddings, the distance between the embedding for a text 
and an image embedding can be used to provide a similarity score. The generative visual process by Fernando et al. \cite{fernando2021generative} uses this text-image similarity as a fitness function to evolve images towards a given text prompt. CLIP can also compute gradients which can be used to provide fine-grained error signals for an image generator. CLIPDraw by Frans et al. \cite{frans2021clipdraw} is an example where images composed of splines are generated to match text prompts. Gradients generated from the text-image mismatch are propagated to the differentiable spline rendering system to progressively optimise their characteristics and placements.

Our work considers the system developed by Latham and Todd \cite{Todd1992}, which has been refined over the recent years, and how it can communicate with one of the latest AI foundation models, Gemini. The next two subsections cover reasons for these two selected systems and their background in more detail.


\subsection{From FormSynth to Organic: FormGrow and Mutator}

``FormSynth'' was first designed in the early 1980's as a rule based system using simple primitive forms (such as cones, cubes, ellipsoids) with transformations or deformations (such as ``bulge'' and ``scoop'') acting on these primitives and their assembly. Initially the artist, William Latham, creator of the generative system, would choose rules to apply, and proceed by drawing the resulting more complex form. Iteratively, this process could be applied for an extensive number of steps, to draw out huge evolutionary trees of forms, some as large as 10 metres, including around 1000 different evolved forms \cite{Latham1989}.



\begin{figure}
    \centering
    \includegraphics[width=0.99\columnwidth]{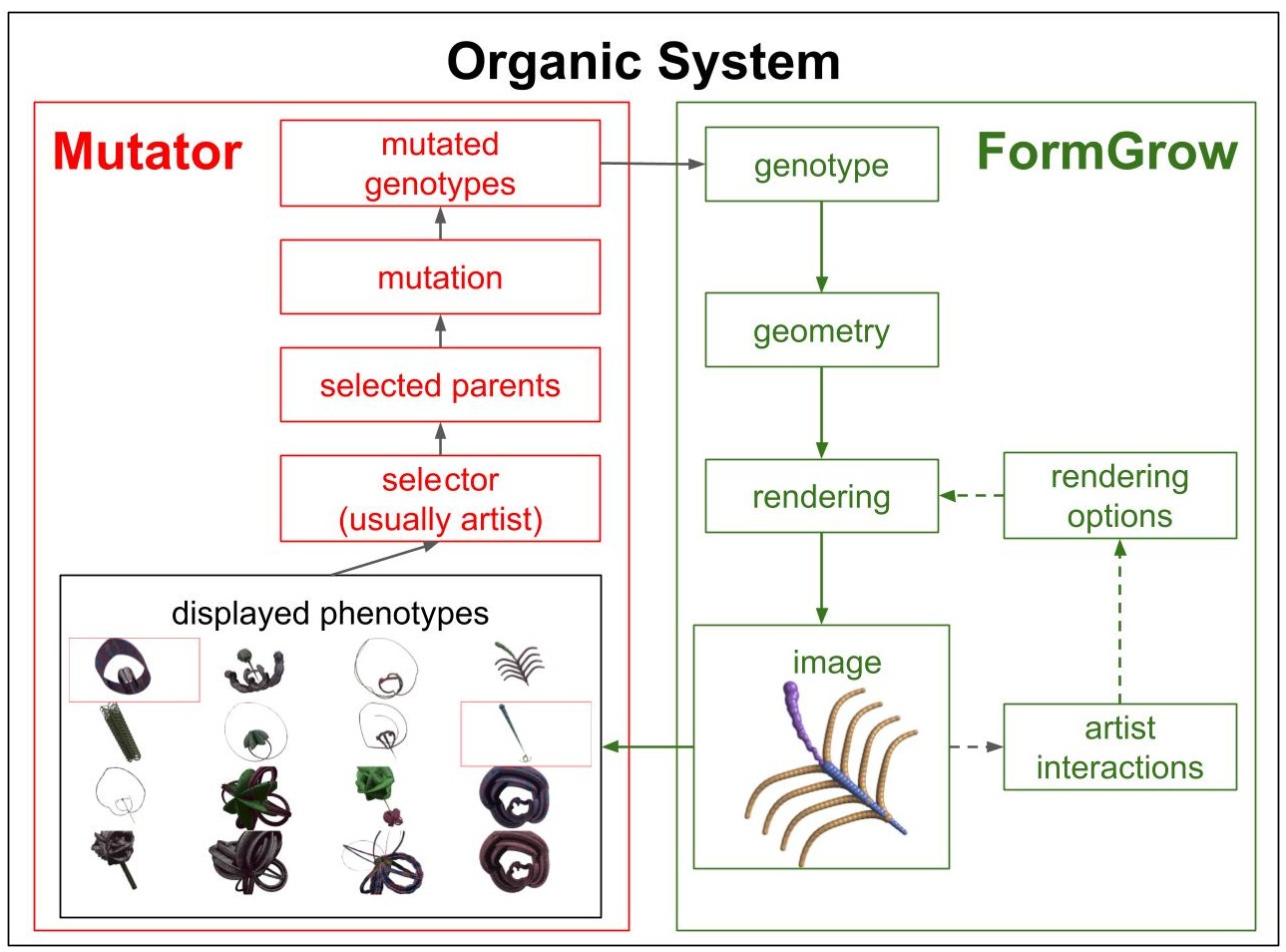}
    \caption{Organic: Joint FormGrow grammar and Mutator genetic algorithm, illustrating the evolutionary loop.}
    \label{fig:organic}
\end{figure}

Latham later began a collaboration in 1987 with mathematician Stephen Todd in the use of computers for art generation.
We have implemented their work into a combined architecture relating FormGrow with Mutator, which we call ``Organic'' (Fig.~\ref{fig:organic}). The Organic system retains two important aspects from the original published design: ``FormGrow'' (1987) for form generation and rendering, and ``Mutator'' (1989) for a computational simulation of the process of evolution (algorithmic details on the systems are available from the book by Todd and Latham \cite{Todd1992}). 
The Organic system includes other features, such as form animation and camera interaction, but these are not used for \Fevol\ described here.

\begin{figure}
    \centering
    \includegraphics[width=0.95\columnwidth]{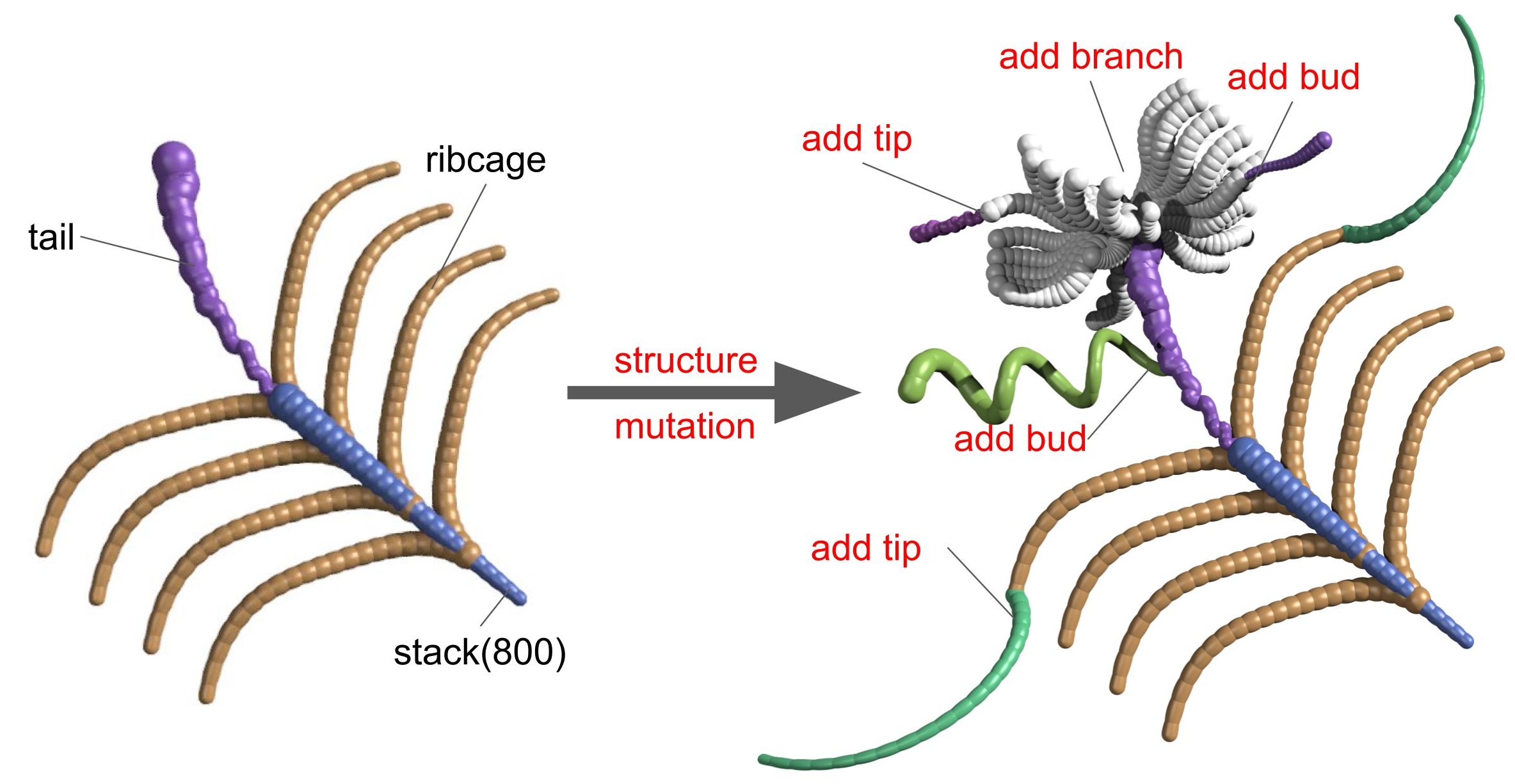}
    \caption{Left: Phenotype resulting from a FormGrow genotype; Right: Result after a structure mutation. Left: form's \textit{tranrule} is:
    horn('QC').ribs(20).radx(40,20).scale(1.2).stack(700).twist(400,{k:36}).bend(67) horn('QT').radx(16,75).scale(0.8).stack(740).bend(24).randxyz({k:20},{v:60}) horn('Q').ribs(20).radx(20,50).stack(800).ribcage('QC', 3, 0.2, .9).tail('QT') mainhorn="Q" .
    }
    \label{fig:combined_evolution}
\end{figure}

FormGrow is driven by a \textit{genotype}, a set of genes that are mainly numeric, and one string gene (the ``tranrule'') that encodes the structure of the form (example in Fig.~\ref{fig:combined_evolution}, left).
The numbers are placeholders and initial values for the genes. Each invocation of FormGrow substitutes values from the genotype, performs the geometry calculation and rendering to produce an image phenotype. Most of the processing is on the GPU, with the genes implemented as uniforms.\footnote{A ``uniform'' is a global variable that remains constant across all vertices or pixels during a single rendering pass.} FormGrow also accepts other inputs such as a view; so the artist can interact directly with a form without any change in genotype.

Mutator works as a conventional artist-driven evolutionary system, mutating the genotypes. It also accepts an optional (external) cost function (phenotype's score) for automatic mutation without artist intervention. One example of such a function is PixelScore (\S~\ref{mutation_and_pixelscore}).
Mutation of the tranrule is called ``structure mutation'' (Fig.~\ref{fig:combined_evolution}, right).
To support structure mutation we have extended FormGrow with a GLSL horn grammar interpreter.

\subsection{Gemini and Foundation models}

The term ``Foundation Model'' refers to a recent paradigm in AI where a single large-scale model is trained on vast quantities of data covering multiple modalities, such as text, audio and images. Such models can subsequently be applied to a wide variety of downstream tasks. Foundation models have broad capabilities, allowing them to be applied to a wide range of tasks that may not have been explicitly included in their training data. This is unlike traditional AI systems that are generally designed for a specific task, e.g. classifying emails as spam.

Most modern foundation models are based on the \textit{transformer} architecture \cite{NIPS2017_3f5ee243}, which uses a ``self-attention'' mechanism to weigh up the importance of different parts of the input data in order to generate its response. For example, in a long sentence it can infer that a pronoun at the end is referring to a noun at the start. Such long-range dependencies are learned from the training data using self-supervised learning. Such an ability is what permits to subsequently generate and complete coherent paragraphs of text or code.

Large language models (LLMs) are a class of transformer-based models trained on a massive corpus of text, such as books and web sites. LLMs are behind the recent explosion in AI chatbot technology. Vision language models (VLMs) are transformers that are trained on images in addition to text - their input being a mixture of data tokens that represent word fragments and image patches.

For the role of the AI aesthetic evaluator, we required a foundation model capable of cross-modal reasoning to evaluate abstract 3D geometries against semantic targets. We selected the Gemini family of models because, unlike previous AI models that relied on combining separate, pre-trained text and image encoders, or contrastive models like CLIP \cite{pmlr-v139-radford21a} that align two separate networks, Gemini is natively multimodal. This means it was trained from the ground up on multiple modalities simultaneously. It was pre-trained on a dataset of text, images, audio, video, and code simultaneously. Because of this native training it can reason across modalities, which was a requirement for the pareidolia tasks in our experiments - such as identifying early, abstract morphological potential in 'digital clay'. Furthermore, Gemini models were also chosen for their robust support for JSON schema enforcement.\footnote{JSON: JavaScript Object Notation, a data-interchange format (json.org).} This allowed us to constrain the output to capture the detailed cross-modal reasoning in strict JSON format. This ensured that the model's assessments (image pros and cons), justification and final choices are reliably reported and extracted across thousands of automated binary tournaments.

\section{Methodology}

We now describe how we chose to integrate Gemini into the creative process with the Organic system using the \Fevol\ approach.

\subsection{Preliminary Investigations with Gemini} \label{preliminary_invest}
At the start of the project we had little knowledge about Gemini's ability to aesthetically assess images, in particular abstract images that would not have been in its training data. We started with preliminary experiments to probe the ability of Gemini\footnote{At the time of this project Gemini 1.5 Flash and Pro were the available models.} to interpret images from FormGrow. When abstract images were presented with the prompt "What does this look like?" we found the responses from Gemini were very promising, with Gemini referencing visual similarities to real-world objects and remarking on aesthetic qualities of the shapes, colours and textures.
Once confirming that Gemini was able to aesthetically assess images, we manually tested its ability to rank images against each other. We presented from 2 to 12 images and prompted Gemini to select the 'best' based on some aesthetic criteria. 
Initial tests involved providing as input the selection prompt:
\begin{quote}
\textit{
    ``These are computer-generated evolutionary art.
    We are interested in which one to select as the parents of the next generation.
    Therefore, we're looking for the ones which have the most interesting aesthetic
    attributes or promise for interesting novelty. Give a detailed argument for each. 
    Then state which you prefer to select as parents for the next generation. 
    In your answer state the number of images you have considered and the numbers of 
    the ones you have selected for the next generation. Format this final decision as
    a comma separated list in square brackets''}
\end{quote}

This prompt was immediately followed by a series of images. We repeated this 20 times, shuffling the images each time. The results exhibited a position bias, with Gemini preferably selecting the earliest images in the context; when using 12 images some of the latter ones appeared to be completely overlooked by Gemini. This is quite likely related to the ``Lost in the Middle'' phenomena \cite{liu-etal-2024-lost} where information in the middle of long text contexts carries less importance in LLM outputs. 
To reduce the position-bias we tried varying numbers of images and using different methods of interleaving the image data with prompt text, as outline in table \ref{tab:context_structures}.
\begin{table*}[t]
\centering
\caption{Comparison of prompt context structures for image selection, where [prompt] is the selection prompt, [image N] is the token data for image N, and ".." is text inserted in the context.}
\label{tab:context_structures}
\begin{tabular}{|l|p{9cm}|p{6cm}|}
\hline
\textbf{Method} & \textbf{Context Structure} & \textbf{Notes} \\ \hline
start & [prompt], [image 1], [image 2], [\dots], [image N] & Task prompt followed by the list of images. \\ \hline
end & [image 1], [image 2], [\dots], [image N], [prompt] & List of images followed by the selection prompt. \\ \hline
topntail & [prompt], [image 1], [image 2], [\dots], [image N], [prompt] & Prompt placed at both ends of image sequence. \\ \hline
interleave\_pi & [prompt], [image 1], [prompt], [image 2], [prompt], [\dots], [prompt], [image N] & Selection prompt interleaved before each image. \\ \hline
interleave\_ip & [image 1], [prompt], [image 2], [prompt], [\dots], [prompt], [image N], [prompt] & Selection prompt interleaved after each image. \\ \hline
tnt\_verbose & [prompt], ``Here are the images:\textbackslash n'', ``Image 1\textbackslash n'', [image 1], ``Image 1\textbackslash n'', \dots,``Image N\textbackslash n'', [image N], ``\textbackslash n That was the last image. Here's a reminder of your instructions.\textbackslash n'', [prompt] & Comprehensive structure with prompts at both ends, explicit image index labels, and descriptive bookending text. \\ \hline
\end{tabular}
\end{table*}

The best context structure we found was \textbf{tnt\_verbose}, i.e. placing the selection prompt at the start and end of the context, preceding each image with text referring to the image index, and descriptive text at the start and end of the image sequence. Although this was the most effective context structure, it still exhibited position bias for any number of images above two. Given this, for the rest of the experiments we used Gemini to use side-by-side image comparisons as the fitness measure with our genetic algorithm, described in Section \ref{the_ga}.

\subsection{Integration of Organic and Gemini}

Mutator’s traditional human-guided evolution presents all the 16 forms on screen at once for the user to select from. Each form has been generated from its unique set of genes. Once selected, the genes of the selected forms are copied and mutated, replacing the forms that were not selected.
To incorporate Gemini in the creative loop we \textit{replace the human curator} with Gemini, using it to select the better image from random pairs of images.

\begin{figure}
    \centering
    \includegraphics[width=0.75\columnwidth]{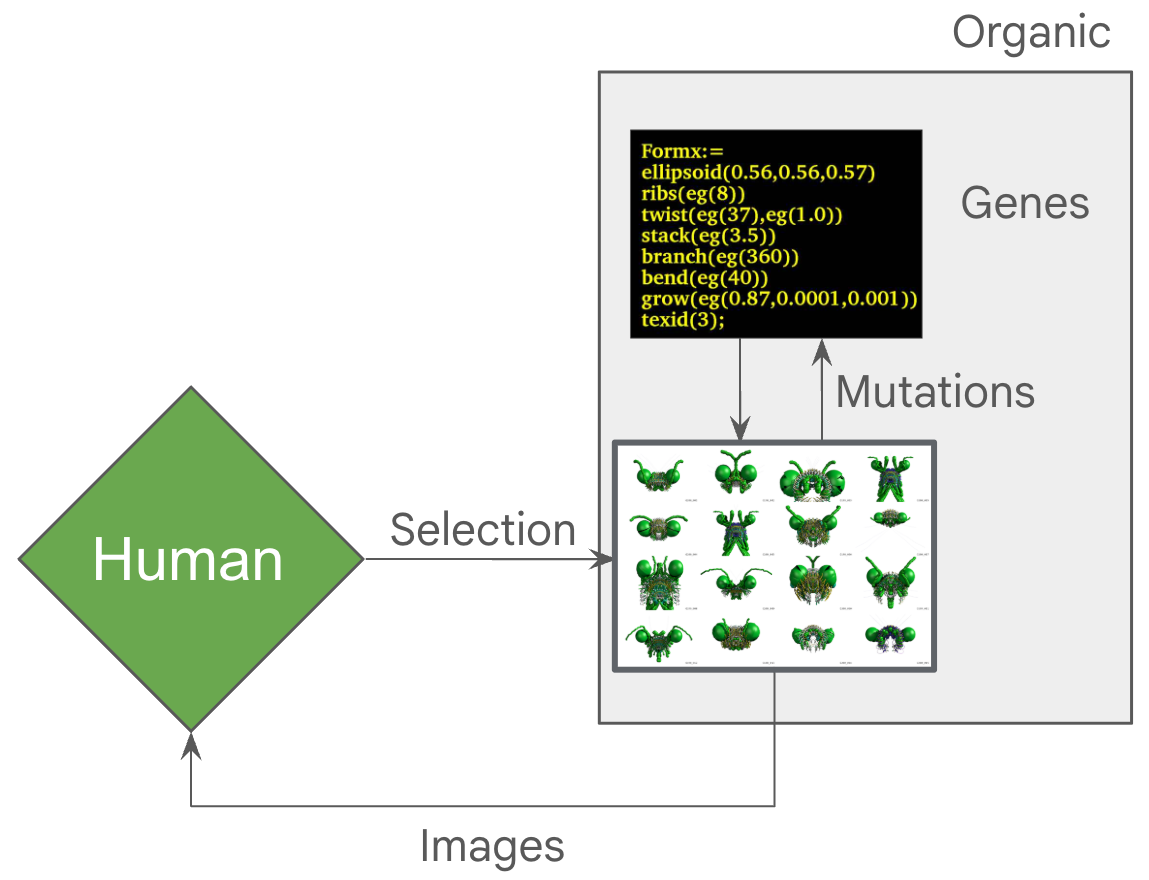}
    \caption{Human operation of Organic to iteratively manually steer the evolution of the form at each generation.}
    \label{fig:human_curator}
\end{figure}

\begin{figure}
    \centering
    \includegraphics[width=0.95\columnwidth]{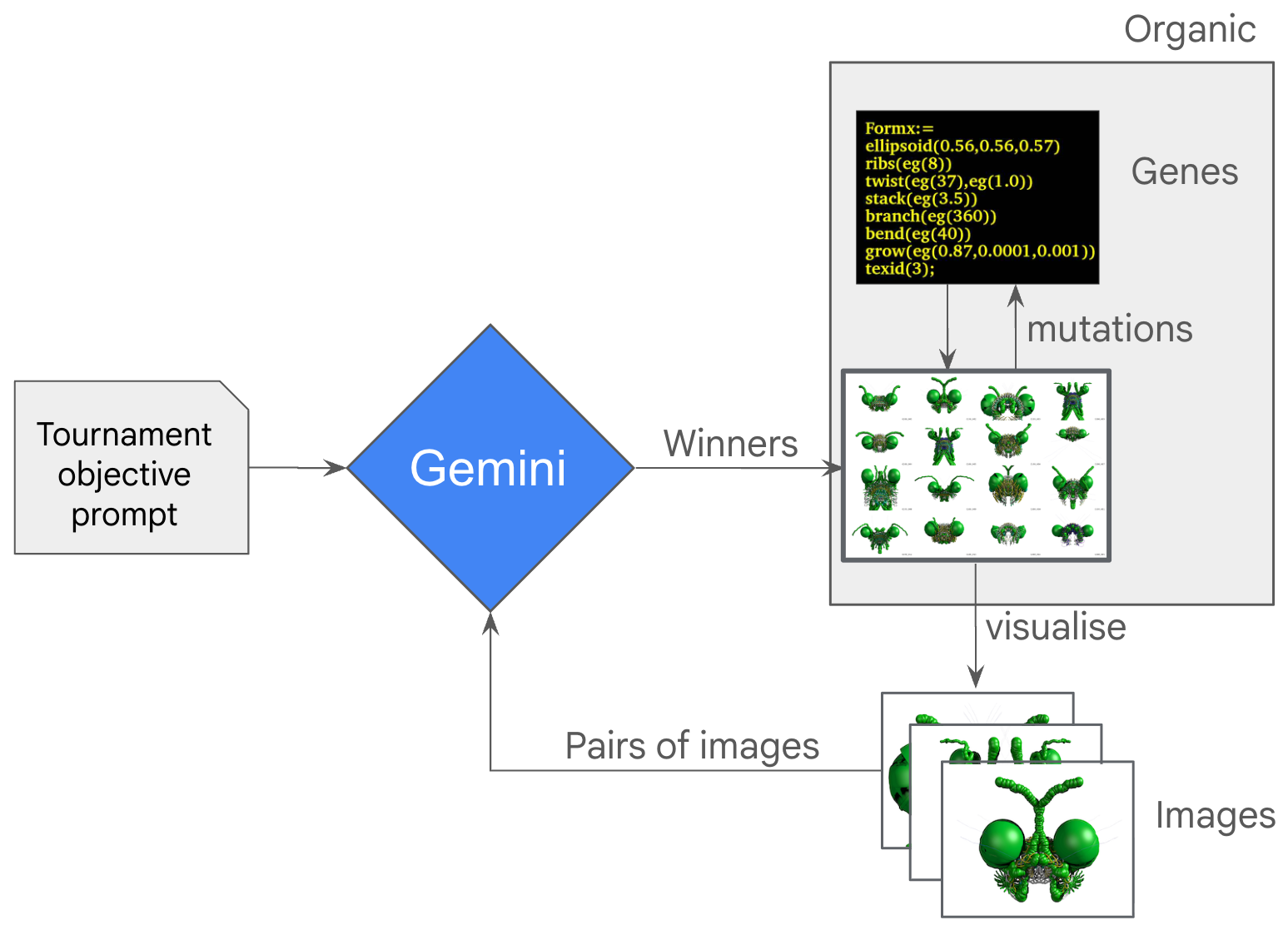}
    \caption{Gemini as the curator, replacing the human in the tedious step-by-step iterative evolutionary explorations.}
    \label{fig:gemini_curator}
\end{figure}

This approach (although limiting in the long term) enabled us to implement a federation of the Organic work with the work on the Gemini side with minimal complexity; keeping a clear separation between both systems. In practice, the interface between the Gemini and Organic is through a shared folder containing files for the current generation (genotypes and their visual phenotypes), an Organic parameter settings file and a file that Gemini uses to write its winning selection for the next generation.

The forms that Organic creates from FormGrow genotypes are 3D structures. Normally, under human-guidance, the user uses the mouse to rotate and zoom each form to assess them from all angles before making their selection. Although such interactive selection is not beyond Gemini’s capabilities, it comes with extra complexity and computational overhead. Considering computational constraints and that we had already confirmed that Gemini was able to aesthetically assess static images, we chose to generate static images from Organic for Gemini to assess. The lighting was fixed and an autoframing mechanism implemented to best frame the wide variety of forms that are generated. Refer to \S~\ref{discussion} for future possible ways of representing the genotypes.

At the start of a generation Organic saves into the folder the genotype and rendered image for each individual in a new population. Once the new generation has been generated Gemini selects the ones to be parents for the next generation (\S~\ref{the_ga} for details) and lists them in a shared file. Gemini also updates a shared parameter settings for Organic, such as mutation rates and whether to use PixelScore (\S~\ref{mutation_and_pixelscore}). Based on these updated instructions Organic applies mutation, renders the mutated forms, and returns a rendered 2D view of the result which is saved into the shared folder. More details of how this interface is used are in Section~\ref{information_flow}.

\subsection{The Genetic Algorithm} \label{the_ga}

\begin{figure}
    \centering
    \includegraphics[width=1.0\columnwidth]{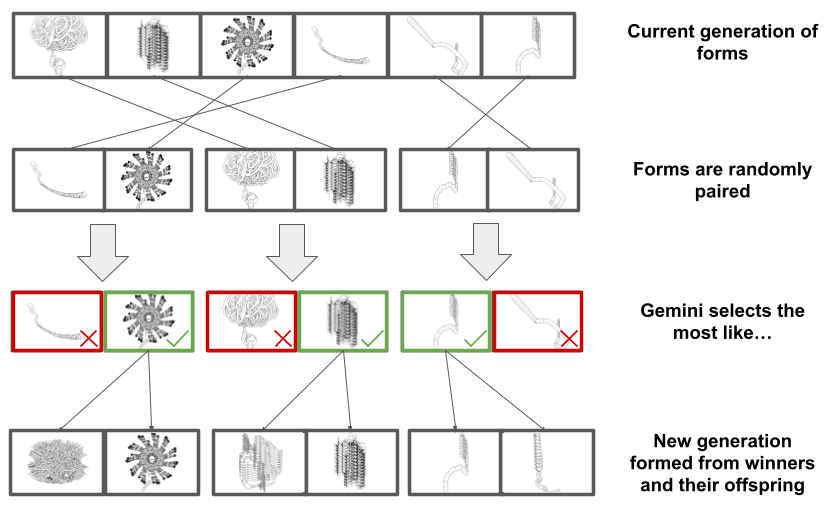}
    \caption{Diagram of the elitist generational genetic algorithm with binary tournament selection.}
    \label{fig:the_ga}
\end{figure}

We incorporated Gemini in the Mutator evolutionary loop by replacing the artist or cost function selection process with a Gemini based process.
For these experiments we implemented an \emph{elitist generational genetic algorithm with binary tournament selection}. The algorithm pseudocode is shown in Algorithm~\ref{algo:the_ga_pseudo} and a diagram of the process in Figure~\ref{fig:the_ga}. We started with a single form and used a fixed population size of 16 individuals. We ran competition between pairs of individuals (\emph{binary tournaments}) for the whole generation to select 8 winners. A mutation was generated from each winner to create 8 ``offspring''. The winners of the generation (\emph{elitist}) together with their offspring go to form the next generation (\emph{generational}). We will now go into the rationale for choosing this particular type of genetic algorithm.

\begin{algorithm}
\caption{Pseudocode: elitist generational core genetic
algo. with binary tournament selection (omitting decreasing
mutation rate \& PixelScore tournaments for readability).}
\begin{algorithmic}
    \STATE $P \gets \text{InitializePopulation}(\text{PopSize})$
    \FOR{$g=0$ to Generations}
    \STATE $\text{EliteSet} \gets 0$, $\text{OffspringSet} \gets 0$\\
    $\text{Pairs} \gets \text{RandomShuffleAndPair}(P)$
    \FOR{$(I_1, I_2)$ \textbf{in} $\text{Pairs}$ }
    \STATE $\text{Phenotype}_1 \gets \text{Render}(I_1)$, $\text{Phenotype}_2 \gets \text{Render}(I_2)$\\
    $\text{Win} \gets \text{BinTourney}(\text{Phenotype}_1, \text{Phenotype}_2, \text{Target})$\\
    $\text{Store}(\text{EliteSet}, \text{Win})$
    \ENDFOR
    \FOR{$\text{Parent}$ \textbf{in} $\text{EliteSet}$}
    \STATE $\text{Child} \gets \text{Mutate(Parent)}$,
    $\text{Store}(\text{OffspringSet}, \text{Child})$\\
    $P \gets \text{EliteSet} \cup \text{OffspringSet}$,
    $\text{UpdateParameters}(P)$
    \ENDFOR
    \ENDFOR
    \RETURN $P$
\end{algorithmic}
\label{algo:the_ga_pseudo}
\end{algorithm}

Binary tournaments were chosen to select winners as the results from the preliminary investigations (\S\ref{preliminary_invest}) indicated that side-by-side comparisons are significantly more reliable for foundation models than attempting to rank or assess a large population (e.g., all 16 individuals) simultaneously. The smaller context size alongside more suitable prompting minimised positional-bias.
One feature of tournament selection is that there is no need to define a global cost function. Each decision is made just based on comparison of the pair of images.

Selecting a generational approach enables us to parallelise the algorithm. The largest bottleneck of the system are the calls to Gemini which can take several seconds per call. Performing all binary tournaments in parallel provided a 8-fold speed improvement for our populations of 16.
We ran each of the experiments for 200 generations. This number of generations was chosen, through experimental trial and error, as the sweet-spot providing sufficient number of generations to enable exploration followed by a refinement, without involving excessive computation time. See \S\ref{App-compute} for a breakdown of computation time and costs.

\subsection{Evolution \& Foundation}
\label{selection}

In natural evolution the chance that an individual survives long enough to have offspring (the individual’s fitness) is a complex interaction of many pressures. In artificial evolution an individual’s fitness is generally a lot simpler, it can vary from a simple metric that needs to be optimised, e.g. minimising the distance a travelling salesperson covers, to a set of interrelated metrics, e.g. the structure of a plane wing with weight/strength/lift trade-offs, or can even be an ill-defined metric, e.g. survival for creatures in a virtual simulation.

A challenge in the field of evolutionary computation has always been \textit{how to specify the fitness criteria} for a given task that can be measured and automated. This is particularly difficult with subjective metrics such as aesthetic assessment. There has been decades of research on computation aesthetics \cite{bo2018computational} attempting to decompose aesthetics to computational metrics, with limited success. Foundation models have been trained on images and language describing them. They therefore provide a unique opportunity to be able to reason over image aesthetics and make judgements grounded in language and semantics. As such they offer a solution to such subjective fitness criteria, opening up the possibilities for evolution based on complex and nuanced competition, and have recently successfully been used to search for interesting forms in 2D artificial life simulations by Kumar et al. \cite{ha-etal-alife-fm}.
However, our approach differs from theirs in three significant aspects:
\begin{enumerate}
    \item \textit{Mechanism of Evaluation}: They use CLIP/DINOv2 to mathematically measure distance, whereas our system uses a native multimodal LLM for active, natural language semantic reasoning via ``chain-of-thought'' prompting.
    \item \textit{Audit Trail}: They treat the model as a silent mathematical fitness function whereas our work treats the generated reasoning and summaries of the AI agent as a core artifact of the artwork itself.
    \item \textit{Application Domain}: They focus on 2D artificial life and open-ended scientific discovery. Our work applies AI to a legacy 3D generative art grammar to sculpt complex organic forms, and is centred on human authorship and co-creative practice.
\end{enumerate}

Our approach is also related to the work of Romero et al. \cite{romero2022evolving}. However, theirs relies on explicit classifiers and training, rather than appropriate prompting of an existing trained generic foundation model. Also, our approach does not need an explicit global cost function, just local (usually pairwise) decisions made by that model.


A benefit of using a foundation model for evaluation is being able to examine the reasoning at every stage. This provides a detailed insight into the decision making and provides a valuable audit trail and narrative to help understand and interpret each run. In the genetic algorithm we checkpoint the state and model outputs for every generation. In Section~\ref{visualisation} we describe how we used this archived data to create interactive visualisations and experiment descriptions and narratives.

\subsection{Prompting Mechanism} \label{prompting}
As outlined in Section \ref{preliminary_invest},
we chose a comprehensive structure with prompts at both ends, explicit image index labels between images, and descriptive bookending text.  
In addition to the context structure we also use a combination of \emph{chain-of-thought prompting} and \emph{constrained token generation} to get the most reliable and consistent results from Gemini. 

\subsubsection{Chain-of-thought prompting}
The binary tournament prompt requires the model to express its aesthetic assessments (pros and cons for each image) and to verbalise its reasoning when making its choice of winning image (e.g., ``this image has a red top-structure like a comb''). This `chain-of-thought'-like prompting ensures that the model's final selection is more reliable, as the choice is grounded in the text it just generated.

\subsubsection{Constrained Token Generation}
We use JSON schema enforcement to constrain the output of Gemini to adhere to a strict JSON format. This ensures that the model's assessments and decision making rationale are explicitly expressed and labelled, and the final choices are reliably extracted from the model's output. 

A significant benefit of the combination of chain-of-thought and JSON-constrained prompting is that we build a complete descriptive and parsable audit trail for each experiment - each of the 1600 binary tournaments\footnote{200 generations, 8 tournaments per generation (for 16 mutations per generation).} are documented in detail with the model's assessment and comparison of each pair of images. This level of transparency provides a unique insight into how the model operates in this domain.

\subsection{The Selection Prompt}

The experiments in this paper focus on target evolution, that is, using a prompt that directs the model towards a specific goal, such as an image that looks like a chicken. Enabling the model to set its own targets throughout the experiment is an ongoing area of research. A typical prompt template for a binary tournament is given in Appendix~\ref{App-prompt}.

\subsection{Gemini/Organic information flow} \label{information_flow}

Our experiments so far have made quite limited exploitation of Organic features. The main experiments limited themselves to a single parent for each new mutation. The interface gives the opportunity for Gemini to change many Organic features dynamically from generation to generation, but to keep initial experiments simple this was only exploited for changing mutation rates. Organic provides features such as the different rendering types (fully rendered and lit, flat shaded, line drawing). These were set up statically by the human experimenters, and many features not used at all. Additionally, Gemini was presented only with the image of a phenotype, and not the genes and structure of the related genotypes.

The use of structure mutation is one case where we broke our ``maximum simplicity'' approach. We initially froze the structure, using a fairly complex fixed structure that we had used in previous experimentations. Though this gives huge variety successful in other contexts, it did not give good results with our more concrete targets from the prompts above. After a time we allowed structure mutation in the experiments, based on the same initial complex structure. This was better, but often where some aspects of the structure gave good results, other aspects could mask these good bits. For the final sequence of experiments we only used a \textit{single stacked horn of spheres} as a starting structure, allowing the structure mutation and selection to evolve complex structures appropriate to the prompt.

As we move forward we expect tighter coupling between the Gemini and Organic code to allow Gemini much greater experimental liberty in the control of the Organic features. For example, we plan to give the Gemini side access to the genotypes so that it can observe what mutations lead to better results. It should then be able to control what mutations to suggest for future generations, and possibly even make appropriate mutations itself.
       
\subsection{Mutation rates and avoiding local minima} \label{mutation_and_pixelscore}

Although our approach is intuitive, simple to implement and efficient to compute, it imposes a very high evolutionary pressure on small populations, risking the evolutionary process getting ``stuck in a rut'' or local minima. In such cases, all individuals become very similar without enough diversity in the population for the process to break free to explore other significantly different possibilities. To overcome this \textit{without human intervention}, we introduce a mechanism for \textit{creative disruption}. Every twenty generations, we automatically switch the tournament goal for two of the tournaments. Instead of asking which image looks more like the goal, a selection is made automatically of the form that displays the most detail and complexity in the image, based on a metric we call \textit{PixelScore}. This enriches the genetic pool and provides the opportunity for the system to explore other evolutionary paths.

\subsubsection{PixelScore}
PixelScore is a simple image processing algorithm aimed at creating diversity in the image. Each pixel is classified depending on which structural component (class) generated it. We compute the coverage for each class, and return the sum of each coverage raised to a power (default 0.2).  The more classes are visible, and the more area is covered by each class, the greater the score. PixelScore can also be used to drive automatic mutation without the use of Gemini. Alternatives of similar image based cost functions have previously been proposed (e.g. refer to \cite{denHeijer2010,Johnson2019}).

\subsubsection{Mutation rate}
A mutation rate controls how much offspring vary from their parents. Experiments start with a high mutation rate which is lowered to almost zero towards the end of the experiment. 
In binary tournaments we inform Gemini what stage it is at in the evolutionary experiment, i.e. we include in the prompt the phrase:
\begin{quote}
    \textit{``Note that the mutation rate used to generate offspring reduces as the experiment progresses. The evolutionary process is currently on step }\{gen\} \textit{out of} \{num\_gens\} \textit{and will therefore stop in} \{num\_gens - gen\} \textit{steps.''}
\end{quote}
 
We found this essential for Gemini to be able to make informed decisions about whether to follow exploratory selections (e.g. early in the experiment where mutation rate is high and there are many more generations to follow) and when it would be better to focus on fine-tuning the design.

\section{Visualisation and Interpretation} \label{visualisation}
When we started our project we were not sure how Gemini would perform at such a task. What does it ‘see’ when it’s given an image? How can it interpret shapes and textures that it has not seen before? Can it relate what it's looking at to text-specified concepts or goals? Can it make aesthetic-based assessments and decisions? 

Initial experiments were very promising, just providing Gemini images, even abstract images, it could describe various aesthetic qualities and make comparisons and analogies to physical world objects. It was interesting to read what it said about the images it is looking at and to see how it is assessing them and what aspects it valued.

\begin{figure}
    \centering
    \includegraphics[width=0.99\columnwidth]{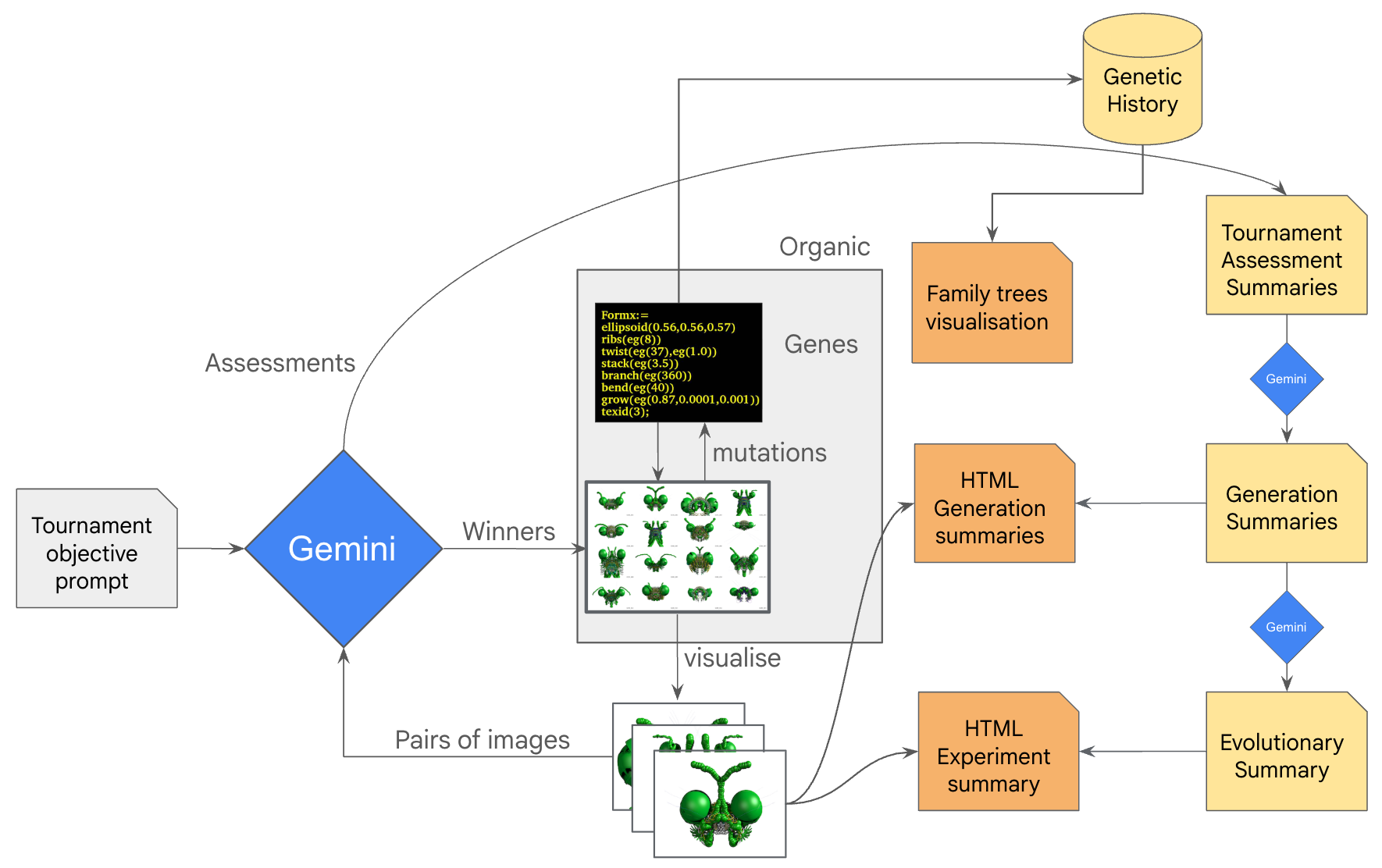}
    \caption{Framework to generate an audit trail for interpretability.}
    \label{fig:interpretation}
\end{figure}

During development it became clear that it was essential to capture all this detailed information during the evolutionary process. This would enable drill down into every decision Gemini makes in an evolutionary run, right down to the reasoned assessments on each pair of images. Figure~\ref{fig:interpretation} details the framework we used to create such audit trails. This level of detail gives a wonderful insight into the \Fevol\ process.

Browsable as web pages are detailed records of each evolution run, with details down to the individual entrant assessments made in all 1600 tournaments. These are generated by Gemini, which is prompted after each generation to summarise the outcome of the tournaments, being asked to identify themes and evolutionary pressures. After the completion of an experiment Gemini can then be prompted to summarise the whole evolutionary run from the generation summaries, identifying evolutionary phases and selecting key moments and individuals.

At the end of each generation we have detailed assessments of each binary tournament. We then prompt Gemini asking it to identify any obvious selection pressures, with justification, and to identify aesthetic trends, e.g. favored features, disfavoured features and any mentioned analogies. Finally we get Gemini to generate a narrative summary for the generation; below is an example:

\begin{quote}
    \textit{Narrative Summary:
    Generation 100 was characterized by an exploratory search for foundational forms that could evolve into a chicken. Selections consistently favored images with a cohesive, rounded, or upright central mass and rudimentary appendages, even if highly abstract. Images that were too chaotic, geometric, plant-like,
    or lacked a central body were rejected. The focus was clearly on establishing a promising structural base with distinct parts, rather than refining details or color.}
\end{quote}

\begin{figure}
    \centering
    \includegraphics[width=0.95\columnwidth]{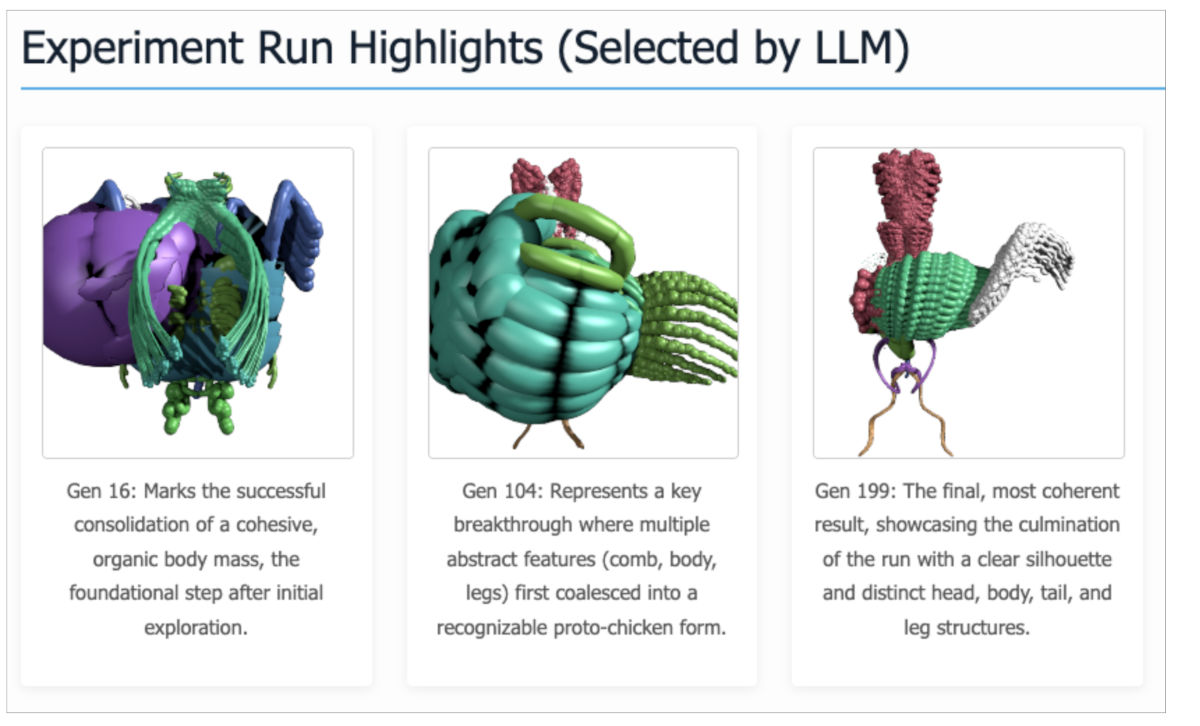}
    \caption{Example of an evolutionary experiment.}
    \label{fig:evol-experiment}
\end{figure}

\begin{figure}
    \centering
    \includegraphics[width=0.99\columnwidth]{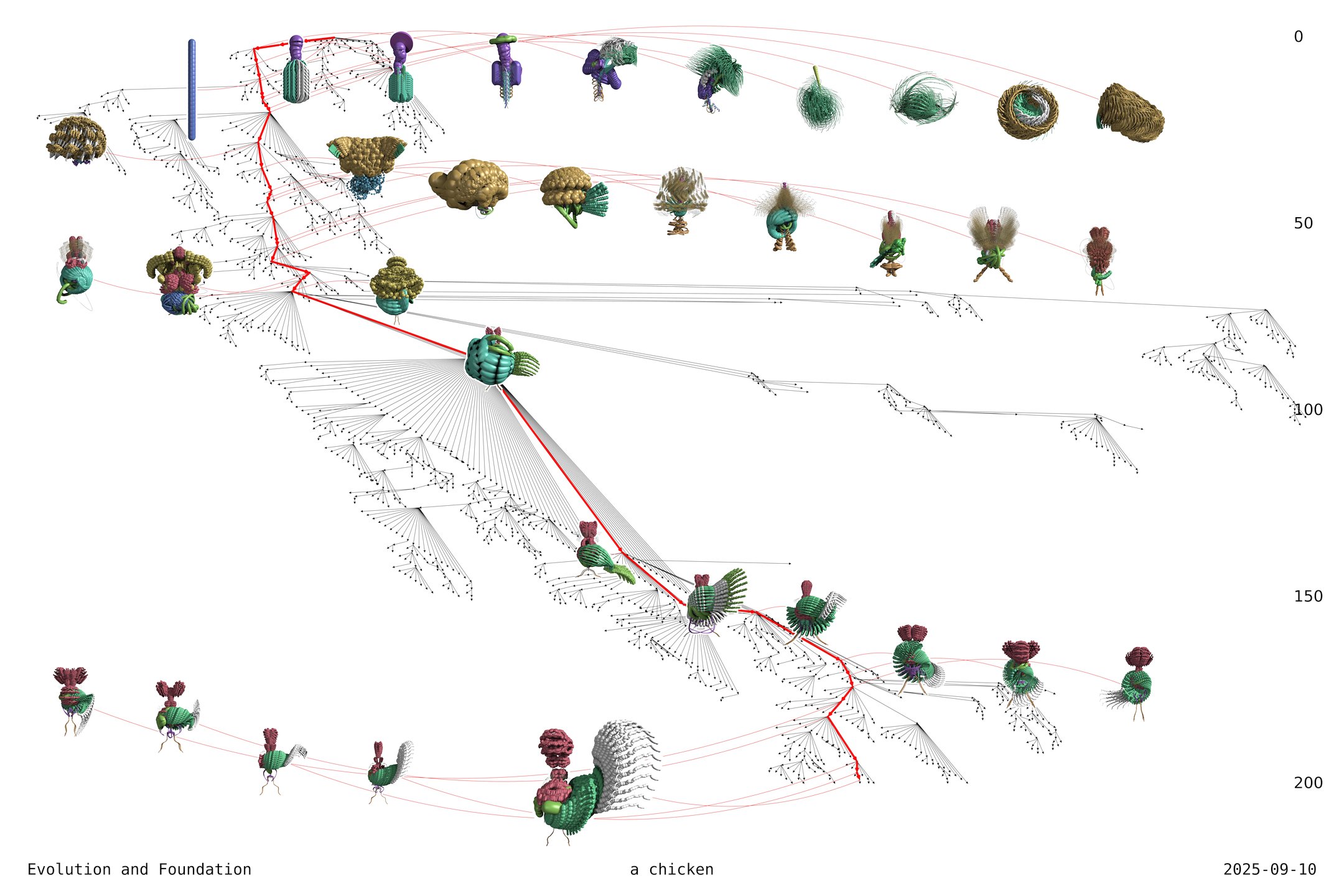}
    \caption{Visualisation of the evolutionary tree produced by Organic and Gemini with goal that of a ``chicken'' form.}
    \label{fig:chickentree}
\end{figure}

Figure~\ref{fig:evol-experiment} gives examples of an evolutionary experiment at various steps (generations 16, 104 and 199), while Figure~\ref{fig:chickentree} gives an overview visualisation of the entire evolution tree with goal that of the form of a ``chicken''.

We finally prompt Gemini to provide a summary of the whole evolutionary experiment using narrative summaries that were created during the run. Gemini generates experiment highlights with a few illustrative images and a summary of the evolution goals, selection pressure phases, lifecycle of key features (e.g. coherent body mass, red comb/head structure) and the evolution of key analogies (e.g. bird-like silhouette, rudimentary legs). As an example, the overall narrative generated by Gemini for the ``chicken'' experiment is given in Appendix~\ref{App-narrative}.

\section{Results}

\subsection{Case Study: The Centric Diatom Evolution}

\begin{figure}
    \centering
    \caption{Visualisation of the evolutionary tree for the ``Centric Diatom'' run.}
    \includegraphics[width=0.99\linewidth]{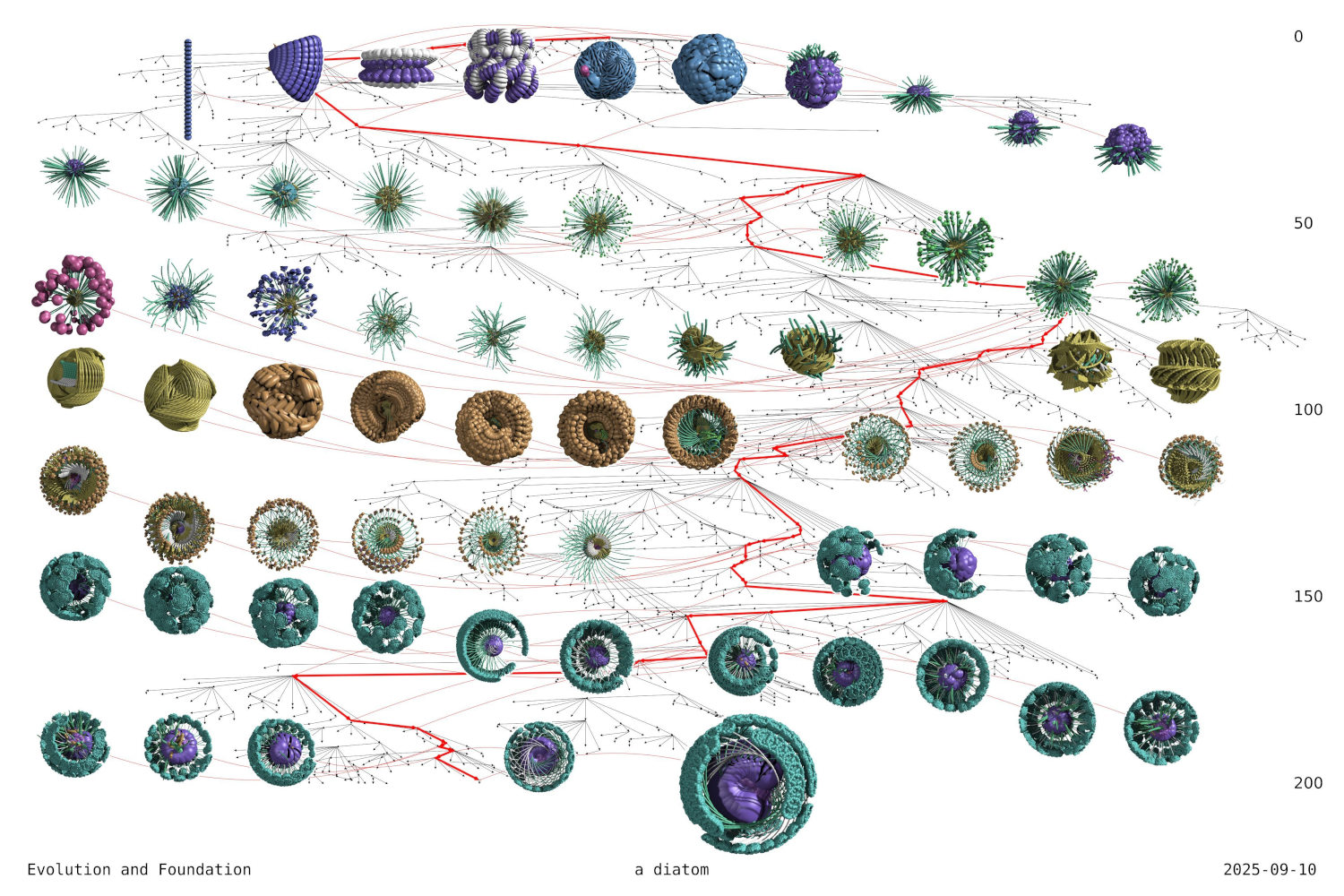}
    \label{fig:diatom_tree}
\end{figure}

Figure \ref{fig:diatom_tree} visualises the phylogenetic trajectory of a candidate from the ``Centric Diatom'' experiment. This lineage represents a successful convergence on the specific biological target: a radially symmetric, microscopic organism with a silica-like shell.

The lineage illustrates three distinct phases of ``Targeted Evolution'' identified by the Gemini aesthetic assessor:
\begin{enumerate}
\item Early Exploration and Symmetry Establishment (Generations 0–12)
The lineage begins with an "Early Exploration" phase dominated by a search for raw evolutionary potential. The AI selector initially filtered for "segmentation and cellular textures" while actively rejecting simplistic or mechanical forms.
\begin{itemize}
  \item Key Turning Point (Generation 4): The lineage highlights a critical early breakthrough where clear radial symmetry and compact, spherical forms emerged. This feature became the "primary structural filter," rejecting chaotic forms to define the evolutionary path toward a "centric" morphology.
\end{itemize}
\item Mid-Run Oscillation and the "Sea Urchin" Trap (Generations 13–91)
The middle section of the lineage displays a "Mid-Run Oscillation" characterized by balanced evolutionary pressure. During this phase, the system navigated a common local minimum where candidates possessed the correct radial symmetry but lacked the specific surface texture of a diatom.
\begin{itemize}
  \item The "Sea Urchin" Analogy: Intermediate forms in this lineage were frequently analogized by Gemini as "sea urchins" or "viruses"—forms that were radially symmetric but too spiky or simple. These served as "stepping stones," selected for their structural foundation but targeted for surface refinement.
\end{itemize}
\item Late-Stage Exploitation and Frustule Refinement (Generations 92–199) 
The final segment of the lineage demonstrates "Late-Stage Exploitation," where selection pressure shifted entirely to surface texture and material quality.
\begin{itemize}
  \item Geometric Organization (Generation 94): The lineage marks a pivotal shift from chaotic complexity to organized geometric patterns, specifically "honeycomb" structures.
  \item The "Frustule" Aesthetic: The final generations (190–199) show the successful evolution of a "glass shell" or "frustule" quality. The AI explicitly selected for "porous" and "silica-like" textures, moving away from organic irregularity toward intricate, manufactured-looking geometry.
\end{itemize}
Final Outcome (Generation 199)
The terminal image in this lineage (Individual 007) represents the highly refined aesthetic of a "centric diatom." It combines the foundational circular form established in Generation 4 with the intricate, porous surface details evolved in the final hundred generations, successfully mimicking the biological target.
\end{enumerate}

\subsection{Further Case Studies}

\begin{figure}
    \centering
    \includegraphics[width=0.85\columnwidth]{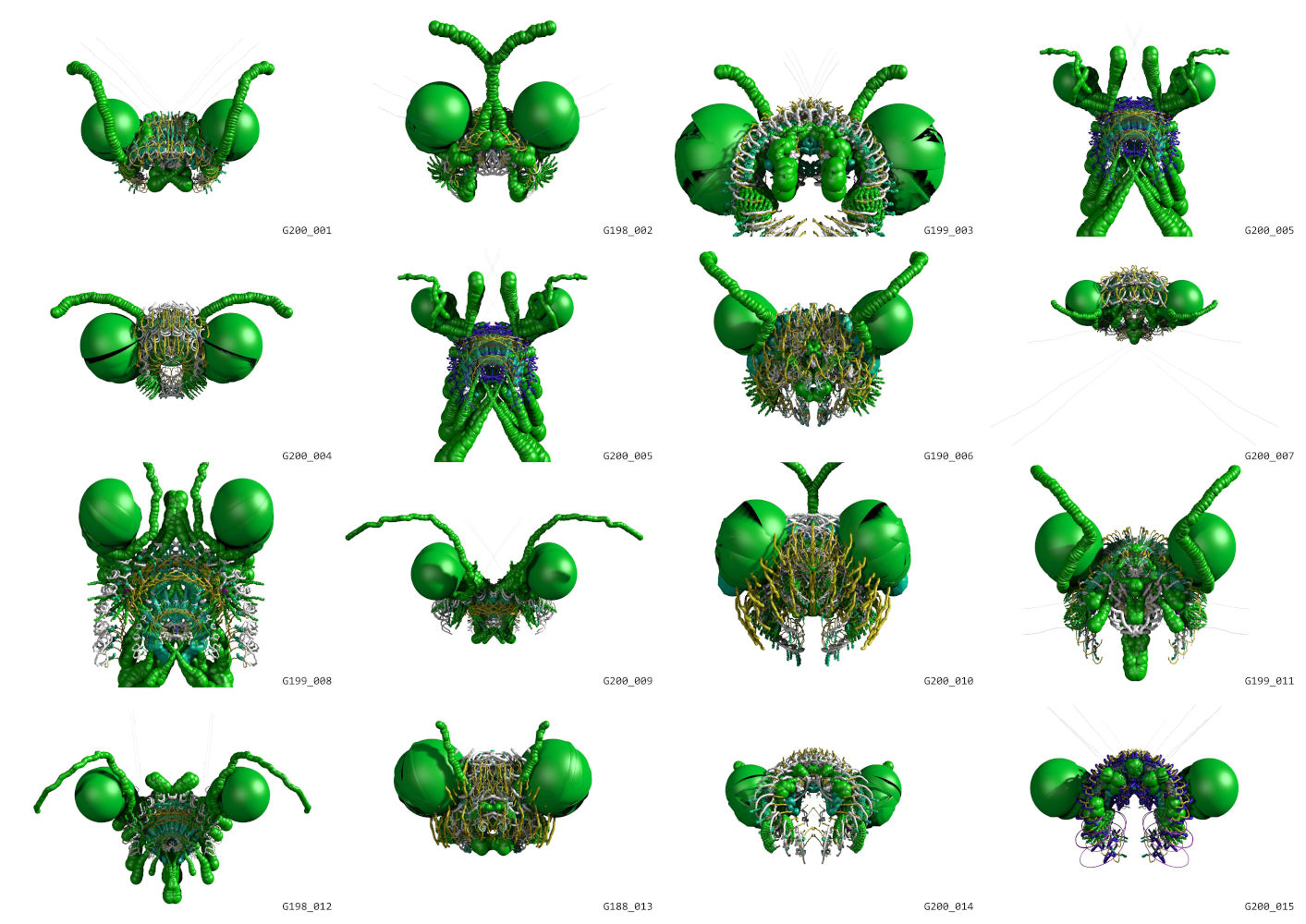}
    \caption{Examples of "fly heads".}
    \label{fig:flyhead}
\end{figure}

The fly heads in Figure~\ref{fig:flyhead} successfully evolved a highly specific biological form - the head of a fly. Gemini focused intensely on anatomical correctness, specifically prioritising compound eyes and antennae.  


\begin{figure}
    \centering
    \includegraphics[width=0.85\columnwidth]{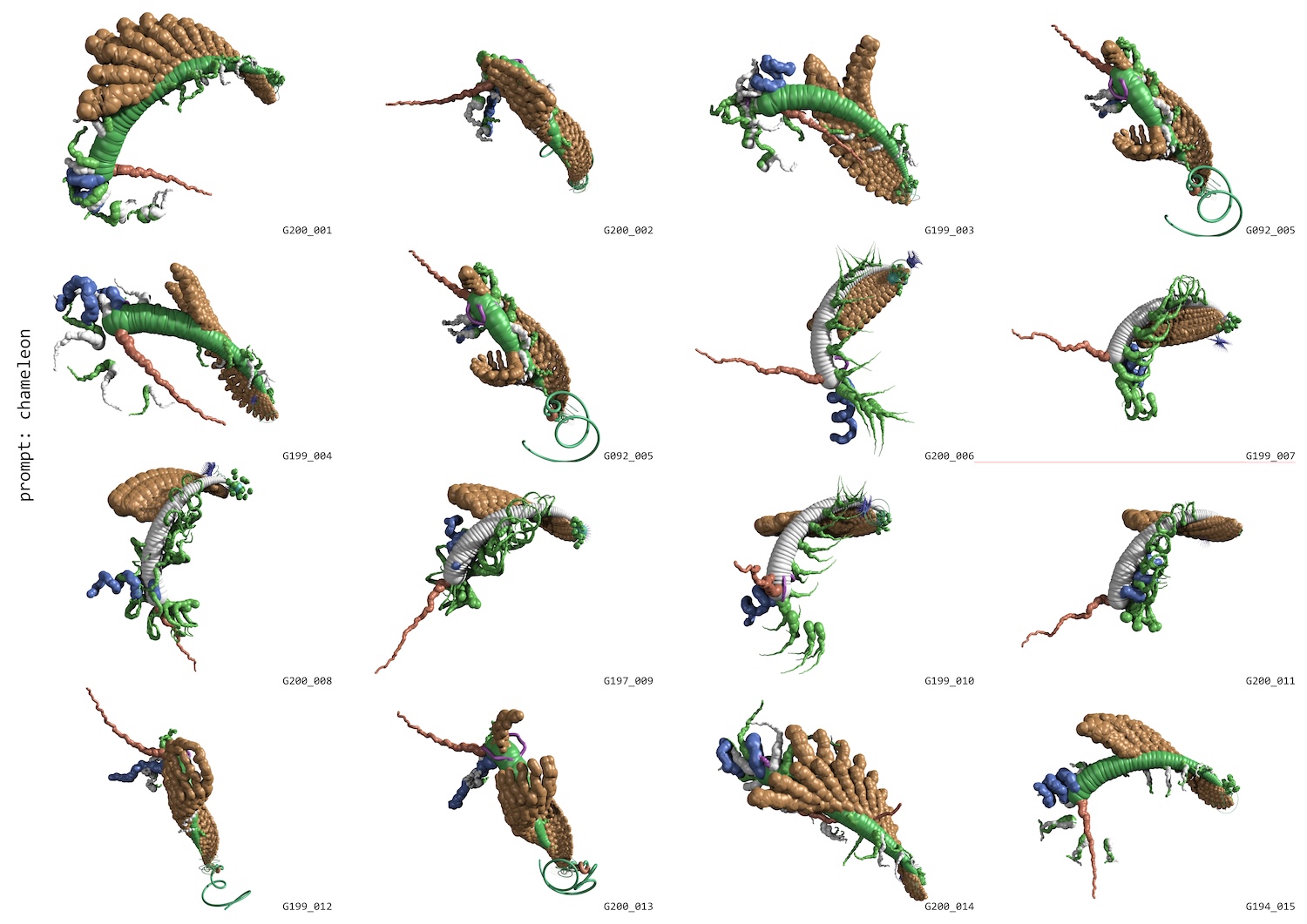}
    \caption{The "chameleon" target was particularly challenging.}
    \label{fig:cham4x4prompt}
\end{figure}

The chameleon was one of the hardest targets to evolve for. Although still highly abstract, the examples of individuals from an evolutionary run in Figure~\ref{fig:cham4x4prompt} show an interesting fusion between chameleon head, tail and branch. 

\begin{figure}
    \centering
    \includegraphics[width=0.95\columnwidth]{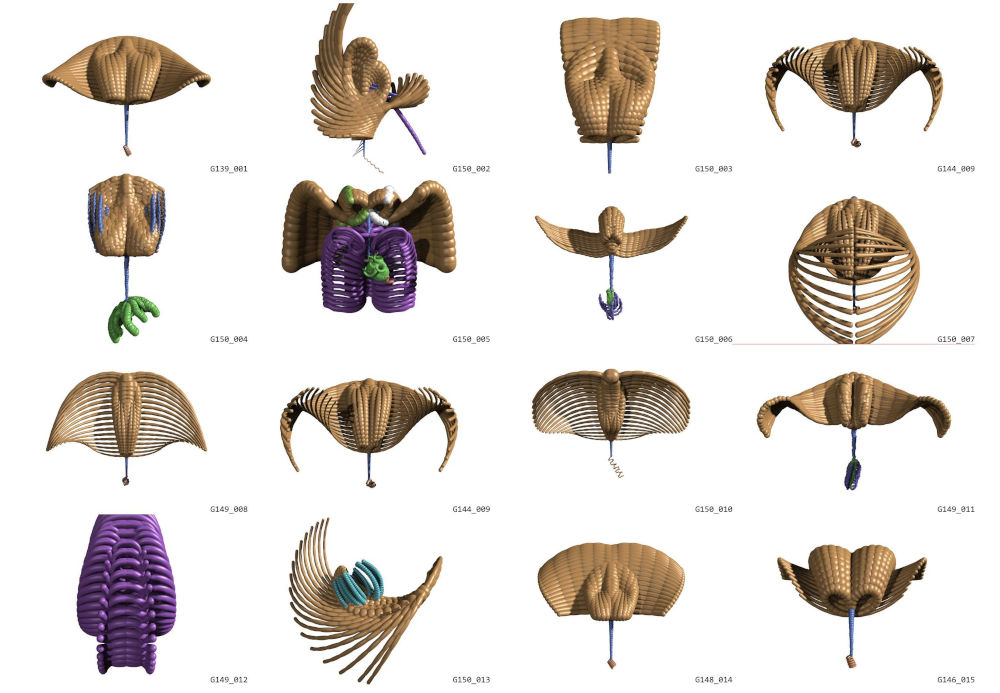}
    \caption{The evolution of a Horseshoe crab, showing a variety of ancestors for a generation 199 individual.}
    \label{fig:crabpair}
\end{figure}

Figure \ref{fig:crabpair} shows an experiment to find a horseshoe crab form, especially the attempts to reach a good symmetry.

\section{Discussion} \label{discussion}

\subsection{Challenges and Limitations}

The process we employed in this work decoupled the creative process into two distinct phases: an initial artist-led phase, selecting the starting genotype, algorithmic parameters, prompt structures, target targets, followed by an autonomous evolutionary loop. This 'set-and-go' approach revealed several critical failure modes, dead-ends and algorithm bottlenecks that highlight the limits of our current framework, persisting despite our use of PixelScore to actively inject genetic diversity (\S~\ref{mutation_and_pixelscore}). 

\subsubsection{Initial Genotype Constraints}

One significant key to unlocking diverse geometric possibilities was to make the initial genotype very simple. In early experiments we attempted to start the process using a fairly complex initial genotype that had been used in previous projects. This starting state had been proven to generate a high variety of structures under human curation, yet it proved limiting when paired with Gemini's target-driven selection.

The complexity of the initial structure meant that potentially positive mutations could be lost in the visually chaotic geometry elsewhere in the rendering. Furthermore, starting with a complex form limits the possibilities of future forms; essentially starting some way down an evolutionary tree; starting with a simple form of stacked horn of spheres provided a ``clean slate'' for the model to sculpt.

\subsubsection{The ``Sea Urchin'' and ``Ammonite'' Attractors}

Across multiple runs in testing we observed that Gemini often guided evolution towards semantic local minima where the forms converged and could not escape to make more progress towards the goal. Two examples were observed:

\begin{enumerate}

  \item The Sea Urchin - in radially symmetric targets, such as the Centric Diatom runs, Gemini quickly converged on the primary structural requirement of radial symmetry. However, once a compact, spiky radial form was achieved, the system often became trapped in sea urchin-like structures with Gemini repeatedly selecting for spiky, high-contrast, multi-segmented individuals because they visually satisfied ``cellular detail''. It struggled to break this structural dominance to select for the delicate, porous, glass-like surface textures required of a true diatom shell.
  
  \item The Ammonite/spiral - FormGrow is particularly adept at generating recursive branching structures and as such the grammar often produces beautiful, tightly curled spirals reminiscent of ammonite fossils. However, Gemini often over-focussed on this biological detail as high-quality composition. Over successive generations this could result with highly organized spirals that could not evolve more complex structures such as legs or heads.
  
\end{enumerate}

\subsection{The Artist as System Designer}

The integration of Gemini into the Mutator workflow represents a significant shift in the role of the human artist. Traditionally, the artist acted as an ``aesthetic gardener'', manually selecting individuals for survival in each generation. In this new framework, the human role transitions to that of a system designer or director. The artist's primary creative input shifts from making individual selective choices to defining the evolutionary ``goals'' via prompt engineering and overseeing the parameters of the digital ecosystem. This allows for the exploration of much larger evolutionary timelines—reaching hundreds of generations—that would be practically impossible for a human to manage manually.

\subsection{Semantic Mapping and Pareidolia} 

The ability of Gemini to perform pareidolia on abstract forms is central to the project's success. The ``chicken'' experiment served as a litmus test for this capability, demonstrating that a multimodal foundation model can find and amplify specific semantic features within a purely abstract, procedural space. For example, in the ``chicken'' experiments there are times when the foundation model selected winners based on characteristics that shown promise to evolve into chicken-like structures in the future, e.g. selecting for a coherent body mass or a red comb/head structure, bird-like silhouette, or rudimentary legs. 

Interestingly, the resulting forms often evoke the ``unnaturalness'' seen in the works of Giuseppe Arcimboldo,\footnote{Italian artist from the 16th century, famous for his paintings of heads made of various objects, such as fruits, fish, books, and more: \url{https://www.giuseppe-arcimboldo.org/}.} where recognizable objects are composed of disparate, organic elements. This aesthetic quality arises from the tension between the ``stubbornly abstract'' nature of the FormGrow grammar and the semantic pressure applied by the AI's goal-oriented selection.

\subsection{Novel Evolution Narratives} 

A unique aspect of this work is the transparency of reasoning enabled by the multimodal LLM. Unlike ``black box'' generative models, the \Fevol\ system provides a detailed audit trail of the AI's ``thoughts'' for every selective decision. By asking the AI to summarize its own evolutionary history, we gain qualitative insights into the phases of the run, such as the transition from ``body mass'' formation to ``structural refinement''. This makes the evolutionary process itself a part of the artwork, moving beyond the final static image.

\subsection{Evolution \& Foundation's Creative Force}

We chose to coin the term \emph{Evolution \& Foundation} to describe the powerful approach of coupling of foundation models with evolutionary processes as we see this as a powerful new approach to exploring generative AI and creativity. 

Evolutionary algorithms have an inherent ability to probe the unexplored through random mutation. However, this alone is insufficient to discover novelty that is interesting. We have demonstrated how foundation models can be used to curate and guide the evolutionary process into areas that are recognisable or have particular aesthetic qualities. It is this balanced tension between the explorative force of mutation and constraining and grounding force of foundation model curation that can give rise to novel and interesting phenomena. 

\subsection{Future Directions} 

Future work will focus on providing Gemini with more comprehensive information and greater agency within the loop. One key area of improvement is the judgment of phenotypes using multiple angle images or 3D video renderings, allowing the model to assess the form's full spatial structure rather than a single 2D projection \cite{Salimbeni2022}. 
Additionally, we plan to explore providing Gemini with greater control over Mutator and FormGrow, and even over its own prompt, extending its creative potential. That will be counterbalanced with improved ability for the artist to take part in the evolutionary loop, see \S\ref{App-artist} for preliminary results.
\section{Conclusion}

In this work, we demonstrated the efficacy of \textit{Evolution and Foundation}, a novel framework that integrates the visual reasoning of large-scale foundation models into
Organic, the Mutator evolutionary process and associated FormGrow form generation system. By this marriage, we have successfully automated the role of \textit{aesthetic selector}, traditionally held by the human artist and more recently by specific aesthetic goal functions \cite{Machado2021}.

Our experiments, particularly the ``chicken'' litmus test, confirm that multimodal models can \textit{perform the digital equivalent of the pareidolia effect common in humans}, identifying and amplifying semantic features within a purely abstract procedural space. This transition shifts the artist’s role from a manual ``gardener'' to a high-level system designer, enabling the exploration of vast evolutionary landscapes that were previously inaccessible.

Furthermore, the system provides an unprecedented level of generative narrative through the model's transparent reasoning. By documenting the AI’s qualitative reasoning for every selective decision and enabling the model to summarize its own evolutionary journey, we have made the process of creation as significant as the resulting artwork. Future work will continue to deepen the AI integration, expanding the AI's agency through multidimensional phenotype assessment and multi-voice interpretive layers.

\appendices
\section{}

 \subsection{Artist in the loop}
 \label{App-artist}

\begin{figure}
    \centering
    \includegraphics[width=0.99\columnwidth]{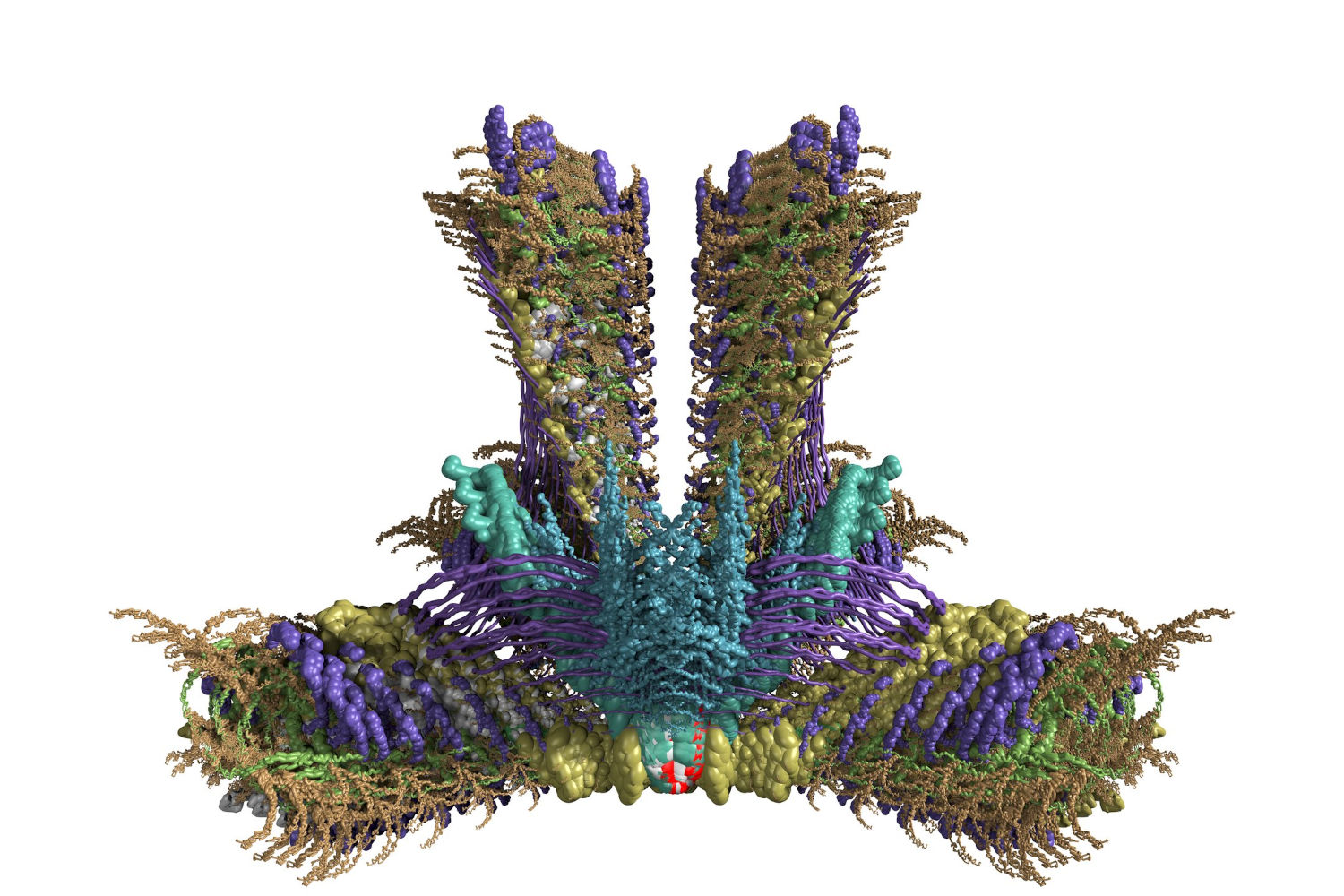}
    \includegraphics[width=0.99\columnwidth]{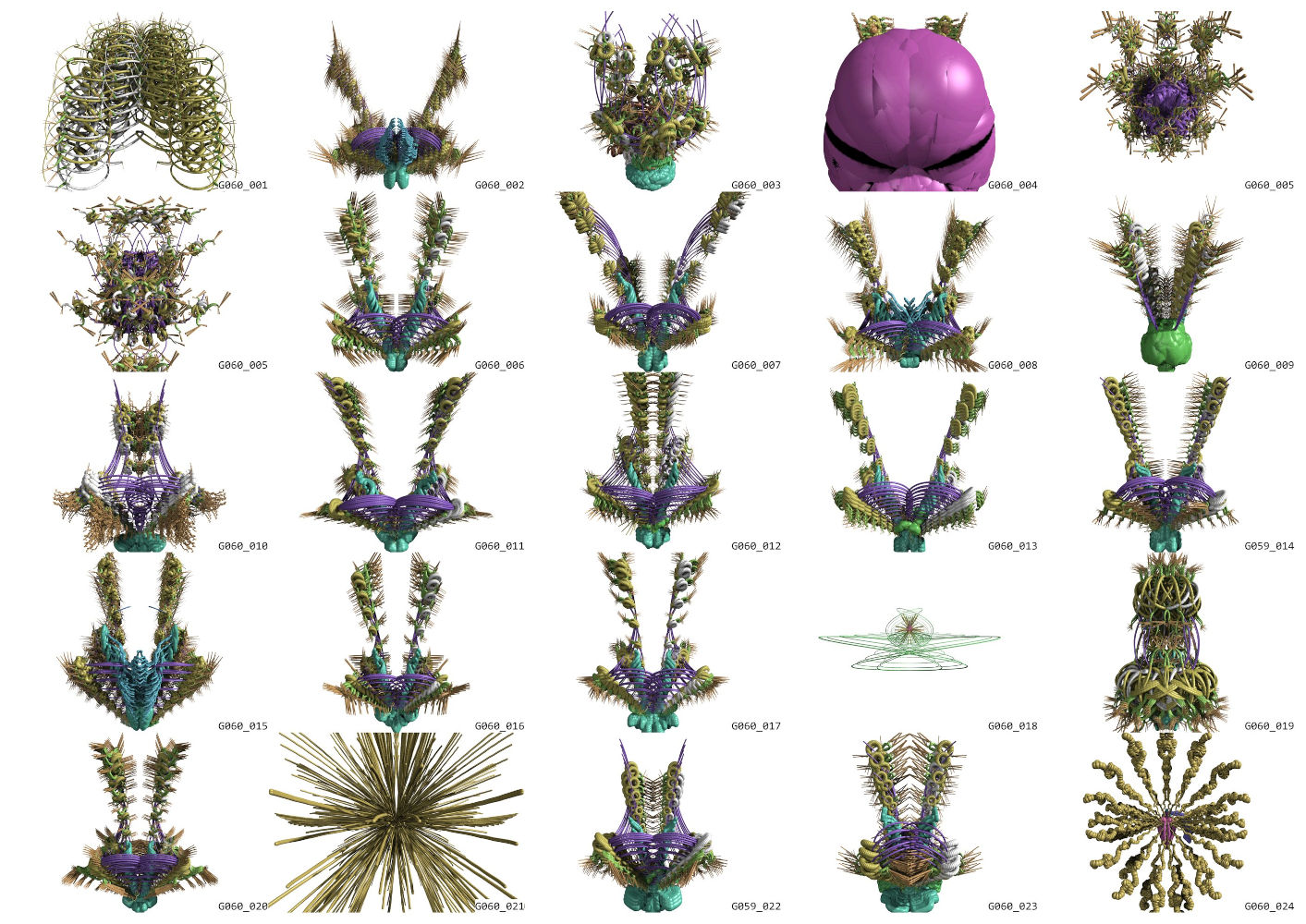}
    \caption{Top: Human selected antler final result. Below: Intermediate set of results at generation 60.}
    \label{fig:antlers}
\end{figure}

As a side experiment we connected Organic to Gemini using more of the traditional Organic mutation/optimization. Mutator prepared a population of 25, and presented them together to Gemini, asking Gemini to score them with a 'looks-like' prompt. The top few (usually 5) were selected as potential parents for the next generation. No extra context was passed; the scoring for each generation used only the 25 images and the prompt. No extra AI information was sought. This simplified interface with only one Gemini round-trip for each generation meant a generation could be created as fast as 10-15 seconds if Gemini was not too busy. This made it feasible to include the artist in the loop; augmenting or overriding Gemini's choices and changing the prompt. Although such an approach will require further experimentations, it led to some impressive --- to the human eye --- results (Figure~\ref{fig:antlers}).

We note that Gemini's scoring during such experiments was very inconsistent from one prompt to the next. Asked for scores in the range [0..1000], one call might give scores in the range up to 500, while an identical call (same images, same prompt) might score almost everything 0, with the highest score in single digits. This is partly related to the ``temperature'' setting of Gemini, which we leave for future work to study in more depth.
\subsection{Selection prompt template}
\label{App-prompt}

A typical prompt template for a binary tournament is shown below, where \_SUBJECT\_ is replaced with the current goal for the experiment, e.g. ``a chicken'' and \_PROGRESS\_ is as described in Section \ref{mutation_and_pixelscore}

\begin{quote}
\textit{ ``You are a critical artist in control of a computer-based evolutionary art process. You are being presented with two randomly selected candidates from the current population of 16. Your goal is to select the best one as a parent for the next generation. You are looking for the image which looks most like} \_SUBJECT\_ \textit{or shows promise to evolve into an image of} \_SUBJECT\_.

\textit{Note that the mutation rate used to generate offspring reduces as the experiment progresses.} \_PROGRESS\_

\textit{Describe your aesthetic assessments and thoughts when making your choice. Use zero-indexing when referring to images by index. Generate a JSON object based on the data provided below:}
\textit{Tournament details:}
\textit{
\begin{itemize}
    \item List of images being compared
    \begin{itemize}
        \item image index (0-based indexing)
        \item positive features
        \item negative features
     \end{itemize}
    \item Index of chosen image (using 0-based indexing)
    \item \raggedright Current goal description: \{current\_goal\_description\}
    \item Is goal finalized: \{is\_goal\_finalized\}
    \item Goal changed this assessment: \{goal\_changed\_this\_assessment\}
    \item Goal change reasoning: \{goal\_change\_reasoning\}
    \item Reasoning narrative: \{reasoning\_narrative\}''
\end{itemize}
}
\end{quote}
\subsection{Example of overall narrative}
\label{App-narrative}

Below is the overall narrative generated by Gemini for the ``chicken'' experiment.

\textit{%
This experiment documents the evolution of a 'chicken' from abstract digital chaos over 200 generations. The journey began with a prolonged exploratory phase (approx. generations 0-20), where selection prioritized any form with 'potential'—typically a cohesive, organic mass—to serve as a foundational building block, rejecting chaotic or overly geometric images. This established a viable 'digital clay' to sculpt.}
    
\textit{The middle phase (approx. generations 21-130) was a long, iterative process of guided convergence. The system moved from sculpting a body mass to establishing a 'bird-like silhouette'. Key features like an upright posture, rudimentary legs, and, crucially, red top-structures interpreted as combs began to emerge and were selectively reinforced. This phase was characterized by the piecemeal assembly of a proto-animal body plan from abstract colored shapes, with many false starts and refinements.}
    
\textit{The final phase (approx. generations 131-199) was marked by strong exploitation and refinement. With a general avian template established, selection pressure focused on improving the clarity and arrangement of parts. A consensus archetype emerged: an upright form with a red 'comb/head', a central body, and appendages for wings, tail, and legs. The last generations were dedicated to fine-tuning this abstract representation, culminating in forms that, while not photorealistic, were unmistakably chicken-like in structure and composition.}

\subsection{Compute Costs and Wall Time}
\label{App-compute}

Each evolutionary run took 3 hours to complete and generated about 5GB of data. Experiments were run on Google Cloud Platform using cost effective VMs based on Intel Haswell-based VMs (4 vCPUs) with 26 GB memory and an NVIDIA T4 GPU for rendering the 3D graphics. Gemini access was through Google’s Vertex AI API. The average cost for an evolution run was \$24, comprising \$16 for Vertex AI and \$8 for VM usage.

\section*{Acknowledgment}
The authors would like to thank Piotr Mirowski, Chrisantha Fernando, Gavin Buttimore, David Warde-Farley, Simon Osindero, Marco Gillies, Jamie Ward, and Miriam Frendo.

\ifCLASSOPTIONcaptionsoff
  \newpage
\fi

\bibliographystyle{IEEEtran}
\bibliography{OrganicAI-new}

\end{document}